\documentclass[fleqn,10pt]{wlscirep}
\usepackage[utf8]{inputenc}
\usepackage[T1]{fontenc}
\usepackage{amssymb}
\usepackage{booktabs}
\usepackage{tabularx}
\usepackage{array}
\usepackage{xcolor}
\usepackage{amsmath,amsthm,mathtools,bm}
\usepackage{graphicx}
\usepackage{multirow}
\usepackage{microtype}
\usepackage{hyperref}
\hypersetup{hidelinks}
\usepackage{cleveref}

\newcolumntype{Y}{>{\raggedright\arraybackslash}X}

\providecommand{\modelname}{\textsc{RD-JEPA}}

\providecommand{\smeanstd}[2]{\ensuremath{#1\pm#2}}
\providecommand{\sbestmeanstd}[2]{\ensuremath{\mathbf{#1}\pm\mathbf{#2}}}

\newcounter{supplementarynote}

\newcounter{supplementarysubsection}[supplementarynote]

\newcommand{\suppnote}[1]{%
  \clearpage
  \refstepcounter{supplementarynote}%
  \section*{#1}%
  \addcontentsline{toc}{section}{#1}}
\newcommand{\suppsubsection}[1]{%
  \refstepcounter{supplementarysubsection}%
  \subsection*{#1}%
  \addcontentsline{toc}{subsection}{#1}}

\title{{RD-JEPA: Predictive latent pretraining for few-trajectory transfer across reaction--diffusion equations}}

\author[1]{Chenhao Si}
\author[1,*]{Ming Yan}
\affil[1]{School of Data Science, The Chinese University of Hong Kong, Shenzhen, Shenzhen, China}
\affil[*]{yanming@cuhk.edu.cn}

\begin{abstract}
Learning surrogates for time-dependent partial differential equations often requires a new simulation corpus when the governing operator changes. We introduce RD-JEPA, a joint-embedding predictive architecture for self-supervised pretraining on reaction–diffusion trajectories. A single model is pretrained on five parameterized systems and then adapted to three held-out systems whose reaction operators and trajectories are excluded from pretraining. Using one, five, or ten complete trajectories from a held-out system, RD-JEPA achieves lower mean relative discrete \(\ell^2\) field error and mean absolute spatial first-difference error than five supervised surrogate baselines, an independently trained control that removes the trajectory-dependent predictive latent pathway, and an architecture-matched model trained from scratch. Within the evaluated equations, output resolution, forecast horizons, and choices of adaptation trajectories, the results indicate that prediction of future-state representations can support data-efficient adaptation across related reaction–diffusion systems.
\end{abstract}
\begin{document}

\flushbottom
\maketitle
%
%
\thispagestyle{empty}

Reaction--diffusion (RD) systems describe how local nonlinear reaction kinetics interact with spatial diffusion to generate organized structures, including Turing patterns, traveling waves, spirals, and self-replicating spots \cite{turing1952chemical}. They provide mathematical models of pigmentation and skin patterning \cite{kondo2010reaction,milinkovitch2023unreasonable}, limb and digit morphogenesis \cite{raspopovic2014digit,sheth2012hox}, engineered multicellular patterning \cite{scholes2019comprehensive,landge2020pattern}, and related non-biological phenomena~\cite{fuseya2021nanoscale}. However, using these models in predictive and repeated-query settings requires repeated numerical integration of stiff, parameter-dependent partial differential equations on fine spatial grids. The resulting computational cost can become substantial in parameter studies, inverse problems, and uncertainty quantification, particularly when the nonlinear reaction term changes and only a small number of trajectories can be generated for the new system.

Learning-based surrogate models can reduce this cost by replacing repeated numerical integration with a trained approximation of the underlying dynamics. Physics-informed neural networks incorporate the governing equations through residual-based objectives \cite{raissi2019physics,karniadakis2021physics}, whereas neural operators learn mappings between function spaces over parameterized families of equations \cite{li2021fourier,lu2021deeponet,kovachki2023neuraloperator,li2024physics}. Graph-based, convolutional, and transformer architectures provide complementary approaches to forecasting spatially distributed physical fields~\cite{sanchez2020learning,brandstetter2022message,gupta2023towards,lippe2023pderefiner,cao2021choose,hao2023gnot,wu2024transolver,holzschuh2025pdetransformer}, and benchmark datasets facilitate comparisons across equations, coefficients, and initial conditions \cite{takamoto2022pdebench,ohana2024the}. Nevertheless, many existing surrogates are trained on a single governing equation or on parameter variations within an equation family with a fixed functional form. Changing the nonlinear reaction term may therefore require a new simulation corpus and equation-specific retraining.  Few-shot and meta-learning methods can reduce the amount of target-system data required for adaptation~\cite{finn2017model,psaros2022meta}, but how information learned from one reaction operator can be reused for another remains less well understood.

Cross-system pretraining provides a possible route beyond equation-specific training. Multiple physics pretraining and Poseidon learn autoregressive surrogates from several physical systems \cite{mccabe2024multiple,herde2024poseidon}, while multi-operator and in-context methods seek to represent multiple governing equations within a shared model~\cite{sun2025towards,yang2023context,alkin2024universal,subramanian2023towards}. In parallel, self-supervised approaches to PDE learning have used masked reconstruction, input-side proxy tasks, and physics-informed contrastive objectives to learn representations that can be reused in downstream tasks~\cite{rahman2024pretraining,chen2024dataefficient,lorsung2024physics}. Recent studies have also begun to investigate joint-embedding predictive architecture (JEPA) representations for physical-parameter inference and physics-informed surrogate pretraining \cite{qu2026representation,yee2026pijepa}. However, evidence remains limited for the use of JEPA-style pretraining in full-field PDE forecasting, particularly when the nonlinear reaction operator of the target system is absent from pretraining. We therefore ask whether predictive latent pretraining can support few-trajectory full-field forecasting when both the target reaction term and all trajectories generated from it are withheld during pretraining.

The JEPA framework provides a natural formulation for this question because it predicts representations of target states from an encoded context rather than directly reconstructing them in the physical field space \cite{lecun2022path}. I-JEPA combines an online context encoder and predictor with a stop-gradient target encoder updated by an exponential moving average \cite{assran2023ijepa}, and V-JEPA extends this principle to video \cite{bardes2024vjepa}. We use reaction--diffusion dynamics as a controlled PDE setting in which the source and target systems share the broad structure of spatial diffusion coupled with local nonlinear kinetics, while differing in their reaction terms. Our hypothesis is that predicting future states in representation space can capture dynamical information shared across related systems and remain useful after the reaction operator changes. RD-JEPA implements this hypothesis through complementary diffusion-inspired and reaction-inspired predictor pathways, which provide inductive biases associated with neighborhood-mediated spatial exchange and pointwise nonlinear kinetics. The model requires no equation identifier, coefficient vector, or symbolic description of the governing equation.

We pretrained a single RD-JEPA model on five reaction--diffusion systems and evaluated the learned representation through a two-stage transfer design. We first examined whether pretraining reduces the amount of equation-specific data required for adaptation on the five source systems. We then adapted the same pretrained model to three held-out systems whose reaction terms and trajectories were completely excluded from pretraining. Comparisons with equation-specific supervised surrogates, an independently trained control without the trajectory-dependent predictive latent pathway, and the same encoder–predictor–decoder architecture trained from random initialization were used to distinguish the contribution of predictive pretraining from that of the downstream architecture or a particular comparator. These experiments address two related questions: whether JEPA-style future-state prediction serves as an effective pretraining objective for PDE dynamics, and whether the resulting representation remains useful when the nonlinear reaction operator changes. In this way, RD-JEPA is evaluated as a reusable representation-learning approach rather than only as an equation-specific forecasting architecture.
\section*{Results}

\subsection*{Predictive latent pretraining targets reusable dynamics}
\label{sec:results_framework}
We investigated whether predictive pretraining across multiple reaction operators could yield a dynamical representation that remains useful even when only a small number of trajectories are available from a target system. We considered multi-component reaction–diffusion fields governed by
\begin{equation}
    \partial_t \mathbf{u}(\mathbf{x},t)
    =
    \mathbf{D}\nabla^2 \mathbf{u}(\mathbf{x},t)
    +
    \mathbf{R}\!\left(
        \mathbf{u}(\mathbf{x},t);
        \boldsymbol{\theta}
    \right),
    \label{eq:rd_general_results}
\end{equation}
where \(\mathbf{D}\) is the diffusion matrix and \(\mathbf{R}\) is the nonlinear reaction operator with parameters \(\boldsymbol{\theta}\).
The principal transfer experiment withheld the complete target systems: neither their reaction operators nor any trajectories generated from them were used during pretraining. Thus, the target-system experiments test transfer beyond variations in coefficients or initial conditions within a fixed equation family.

\begin{figure}[!htb]
\centering
    \begin{minipage}{\textwidth}
     \centering
     \includegraphics[width=\linewidth]{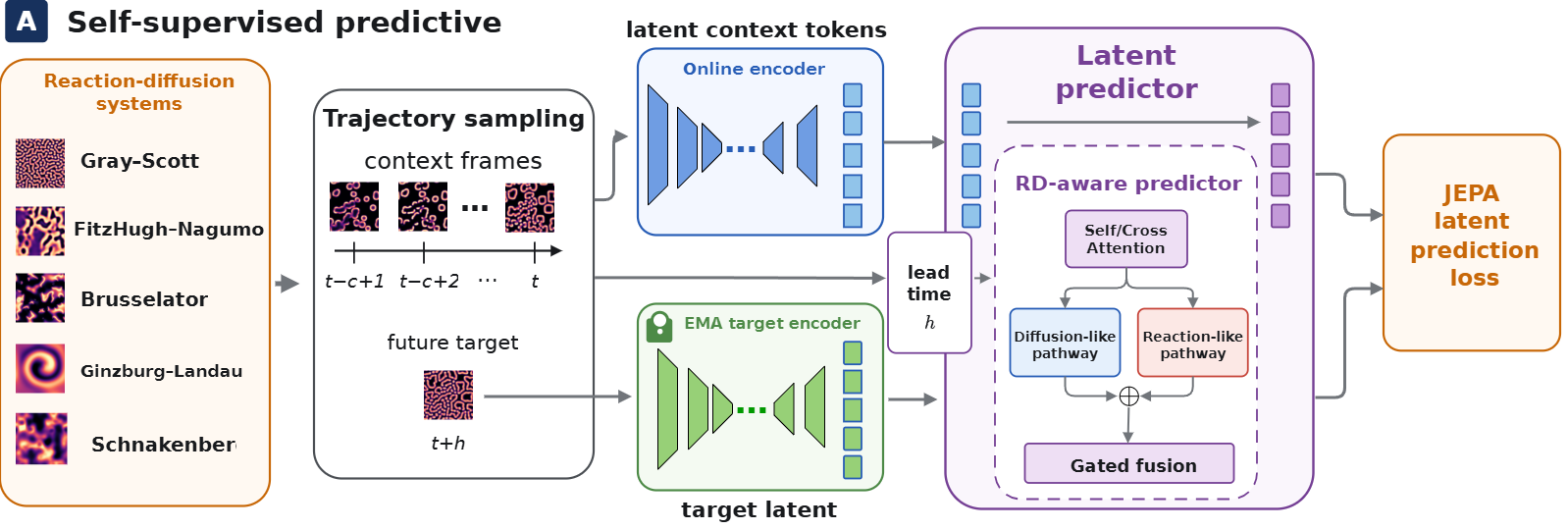}
   \end{minipage}
   
    \begin{minipage}{\textwidth}
     \centering
     \includegraphics[width=\linewidth]{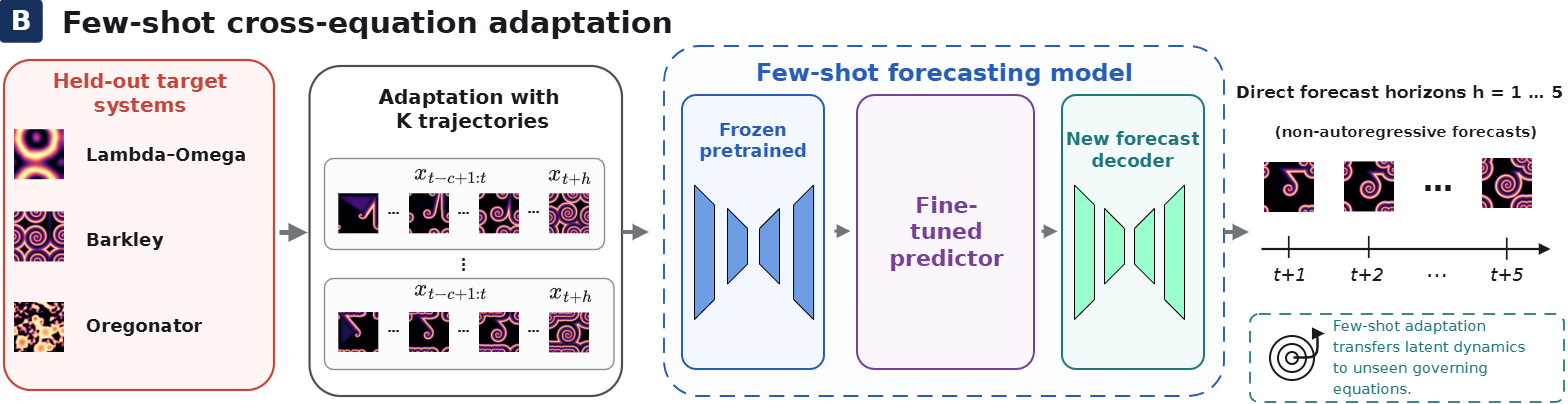}
   \end{minipage}
\caption{
\textbf{a}, During pretraining, a context encoder represents four observed fields and an exponential-moving-average target encoder represents a future field. A lead-time-conditioned predictor estimates the target representation from context; the loss is evaluated in latent space without reconstructing the future field.
\textbf{b}, For downstream forecasting, the source-pretrained encoder is frozen and the predictor is adapted jointly with a dense decoder using $K$ target-system trajectories. The same checkpoint is evaluated on source equations and on governing equations excluded from pretraining, and it produces direct forecasts at $h=1,\ldots,5$.
}
\label{fig:overview}
\end{figure}

RD-JEPA separates self-supervised representation learning from supervised full-field forecasting~(Fig.~\ref{fig:overview}). During pretraining, an online encoder maps four observed fields to a latent representation of the trajectory context, while an exponential-moving-average target encoder represents a future field. A lead-time-conditioned predictor estimates the future target representation without reconstructing the corresponding field. The predictor contains two structured latent pathways: a diffusion-inspired pathway based on nearest-neighbor interactions and a reaction-inspired pathway based on pointwise nonlinear updates. These pathways provide architectural inductive biases rather than numerical discretizations of the physical diffusion and reaction operators. During downstream adaptation, the target encoder is removed, the pretrained online encoder is frozen, and the predictor is fine-tuned jointly with a newly initialized full-field decoder.

A single RD-JEPA checkpoint was pretrained on Gray–Scott, FitzHugh–Nagumo, Brusselator, complex Ginzburg–Landau, and Schnakenberg trajectories. We first evaluated reuse on these five source systems using $K\in\{5,10,20\}$ complete adaptation trajectories. We then adapted the same checkpoint to Lambda--Omega, Barkley, and Oregonator using $K\in\{1,5,10\}$ target-system trajectories. All downstream models received four consecutive states and directly predicted the five future states at horizons $h=1,\ldots,5$, where \(h\) is measured in stored-frame intervals rather than through autoregressive rollout. Each trained model was evaluated on the same 300 held-out test trajectories for the corresponding equation. Support-set construction, preprocessing, comparison-model configurations, optimization settings, and downstream objectives are detailed in Supplementary Note~\ref{supp:note_dataset}--\ref{supp:note_experiments} and Supplementary Table~\ref{tab:supp_numerics}--\ref{tab:supp_baseline_optimization}.


For each equation and adaptation budget, we evaluated three prespecified adaptation-set selections. Within each selection, all models received the same adaptation trajectories and were evaluated on the same test trajectories. Across the three selections, the predictive-pretraining checkpoint, model-training randomness, and test data were fixed; only the selected adaptation trajectories changed. The reported error bars therefore quantify sensitivity to adaptation-set selection rather than variability across pretraining runs, optimization seeds, or test trajectories. For the source systems, the 300-trajectory test set contained 100 in-distribution trajectories, 100 coefficient-out-of-distribution trajectories, and 100 initial-condition-out-of-distribution trajectories.

We report the relative discrete \(\ell^2\) field error and the mean absolute spatial first-difference error. The first measures the global discrepancy between the predicted and reference fields, whereas the second measures mismatch in their local grid-scale spatial variation. Values shown in the main figures are the mean and sample standard deviation across the three adaptation-set-level means. Because no formal hypothesis tests were performed, the comparisons below refer to the ordering of the reported numerical means. Split-resolved source-system summaries are reported in Supplementary Table ~\ref{tab:supp_source_split_summary} and the corresponding run-level values are provided in Supplementary Data~\ref{supp:note_dataset}.


\subsection*{Predictive-latent pretraining improves source-system forecasting with limited adaptation data}
\label{sec:results_source}

We first evaluated whether the pretrained representation improved forecasting on the five systems that contributed trajectories to pretraining. For each source equation and adaptation budget, RD-JEPA was compared with an equation-specific Fourier neural operator (FNO) trained from random initialization and an independently trained no-predictive-latent control. The control removes the JEPA encoder–predictor pathway but retains the full-field decoder, the full-resolution pathway from the latest observed state, and forecast-horizon conditioning. Because all five equations were represented during pretraining, this experiment evaluates data-efficient reuse on source systems rather than transfer to a previously unseen reaction operator.

\begin{figure}[!htb]
\centering
    \begin{minipage}{\textwidth}
     \centering
     \includegraphics[width=\linewidth]{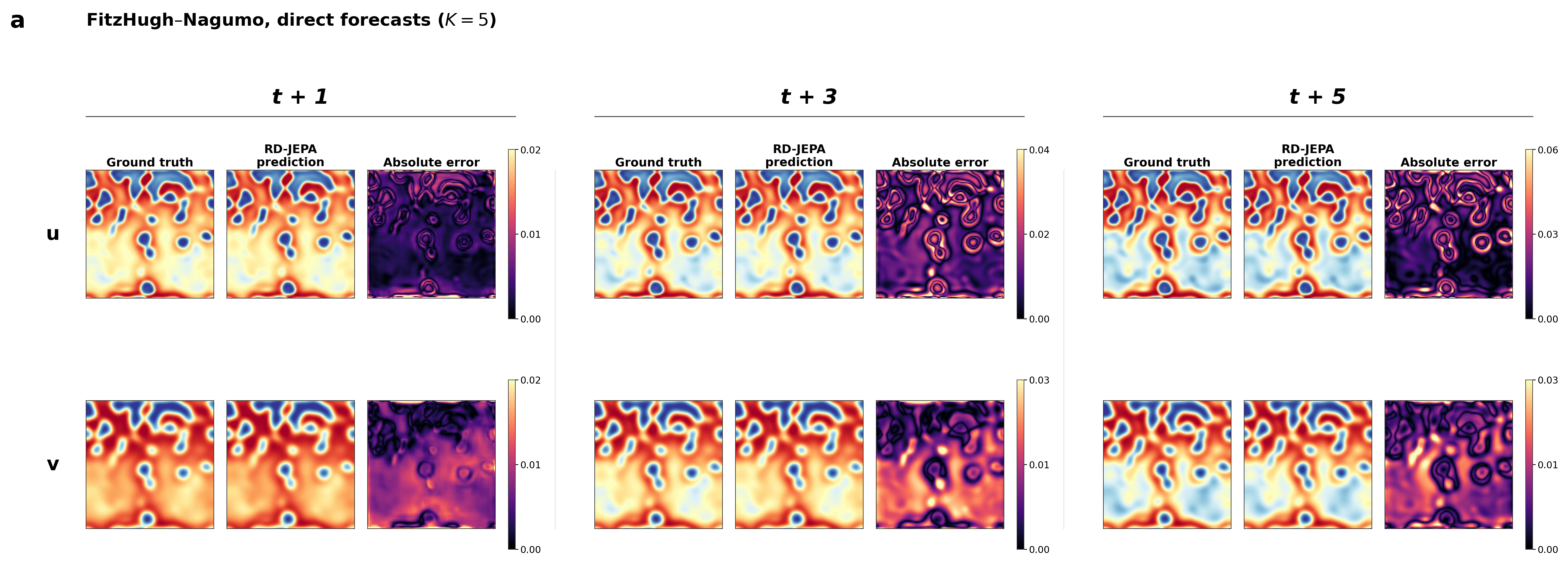}
   \end{minipage}

\begin{minipage}{0.44\textwidth}
     \centering
     \includegraphics[width=\linewidth]{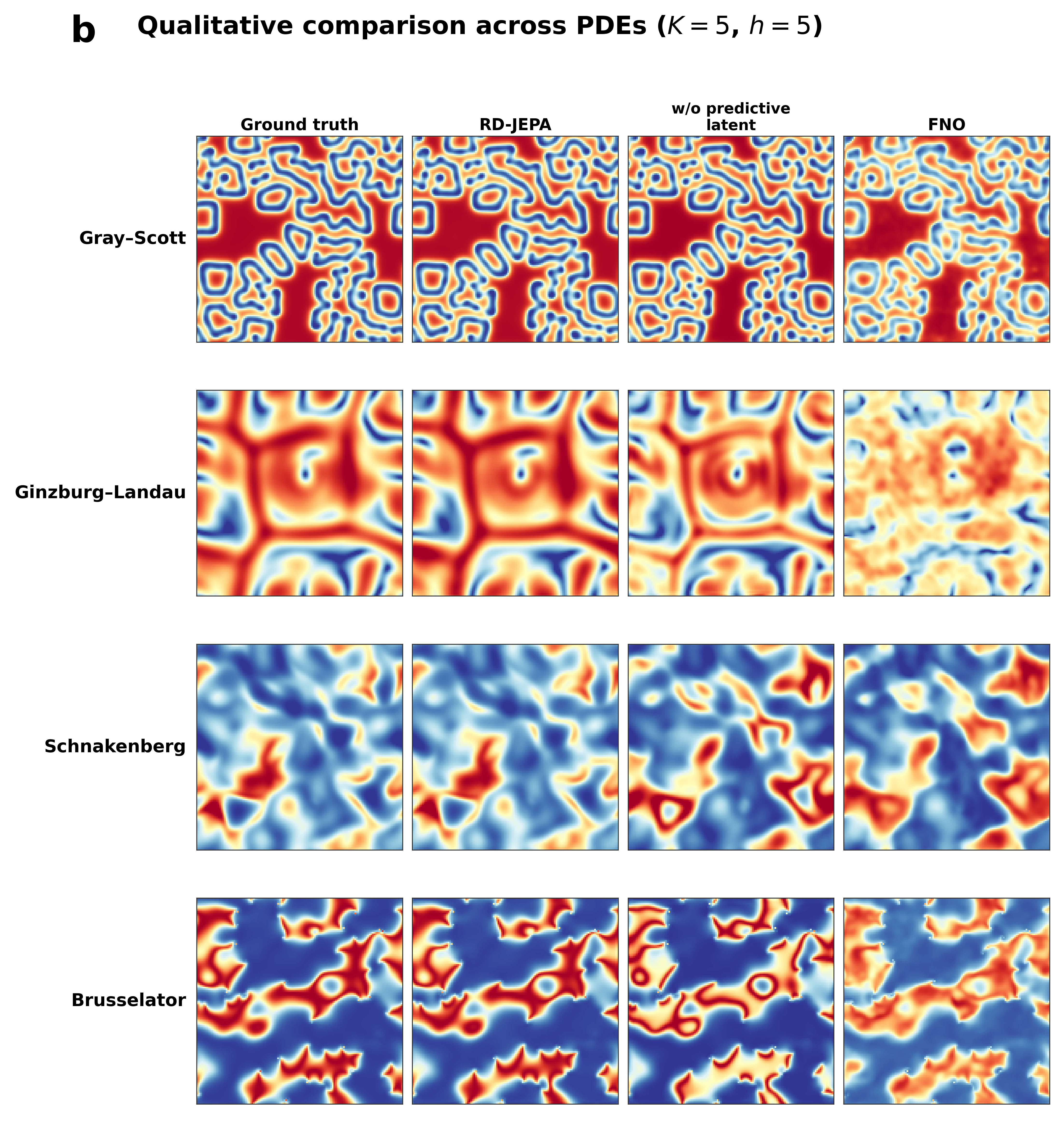}
   \end{minipage}
    \begin{minipage}{0.55\textwidth}
     \centering
     \includegraphics[width=\linewidth]{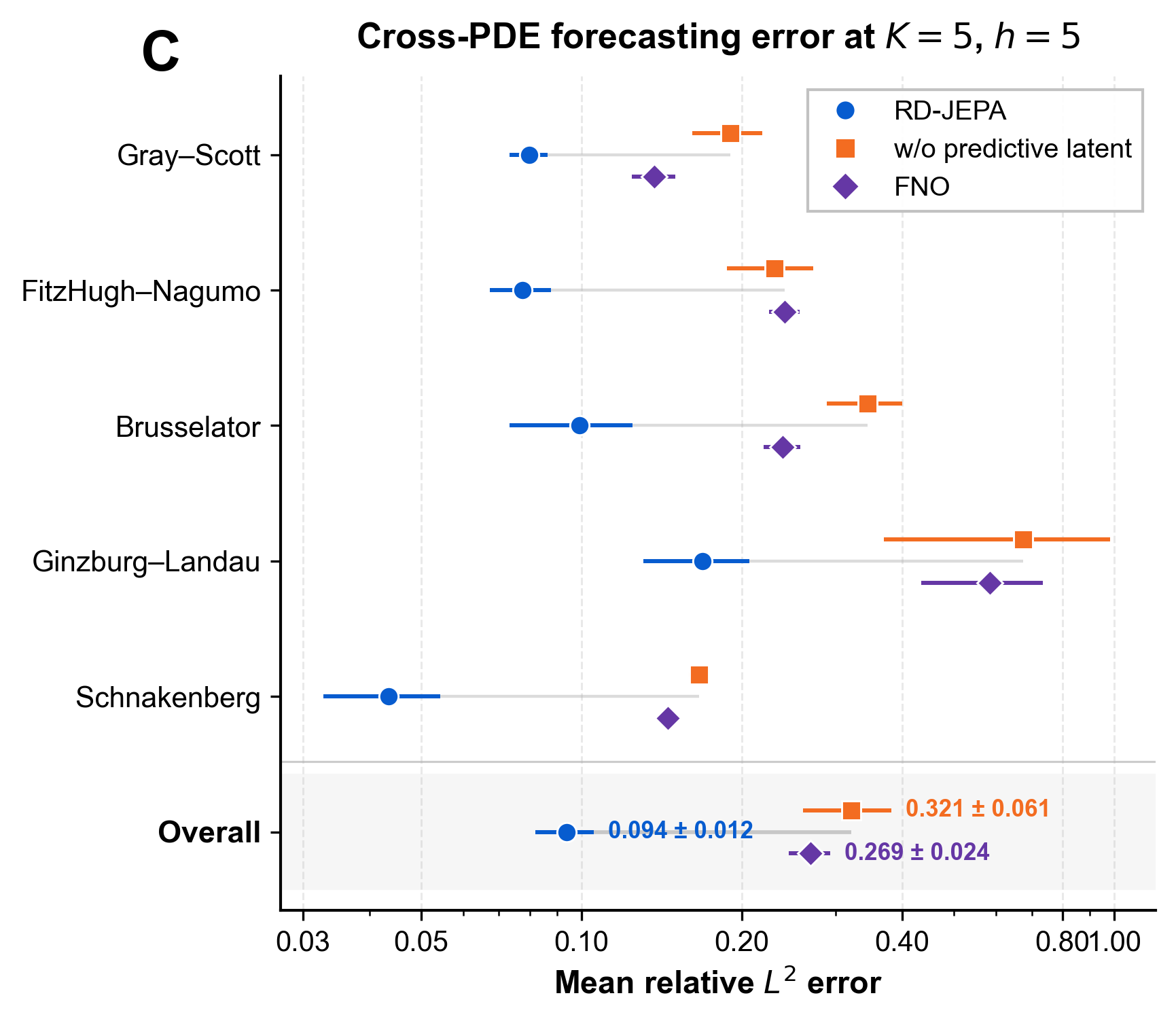}
   \end{minipage}
\caption{
\textbf{Few-trajectory forecasting on the source reaction--diffusion systems.}
{\textbf{a}, Direct forecasts from the prespecified representative downstream run for a FitzHugh--Nagumo test trajectory after adaptation with \(K=5\) trajectories. The two field components, \(u\) and \(v\), are shown at forecast horizons \(h=1\), \(3\) and \(5\). For each component and horizon, columns show the reference field, the RD-JEPA prediction and the pointwise absolute error. Absolute-error colour limits are rescaled independently at each horizon, so magnitudes should be read from the corresponding colour bars rather than compared by colour intensity. The displayed trajectory was selected randomly from the held-out test set.}
{\textbf{b}, Cross-equation qualitative comparison from the same representative run at $K=5$ and $h=5$ for Gray--Scott, Ginzburg--Landau, Schnakenberg and Brusselator. Columns show the reference solution, RD-JEPA, the no-predictive-latent control and an equation-specific FNO trained from scratch using the same adaptation budget. Fields within each row share a common colour scale.}
{\textbf{c}, Mean relative \(L^2\) error at \(K=5\) and \(h=5\), reported separately for each source equation and as an equal-weight mean across the five equations (\emph{Overall}). Each run-level equation score is averaged over the same \(300\) held-out test trajectories. Markers and error bars show the mean and sample standard deviation across three paired support-selection runs, respectively. The \emph{Overall} score is first averaged across equations within each run. Lower values indicate better performance, and the horizontal axis is logarithmic.}
}
\label{fig:source_qualitative}
\end{figure}

At the smallest adaptation budget, \(K=5\), and the longest horizon, \(h=5\), RD-JEPA had the smallest reported mean relative discrete \(\ell^2\) field error on each of the five equations and on their equal-weight average (Fig.~\ref{fig:source_qualitative}c). The numerical ordering relative to both FNO and the no-predictive-latent control indicates that the source-pretrained representation remained useful when only five equation-specific trajectories were available for adaptation.

\begin{figure}[!htb]
\centering
    \begin{minipage}{\textwidth}
     \centering
     \includegraphics[width=\linewidth]{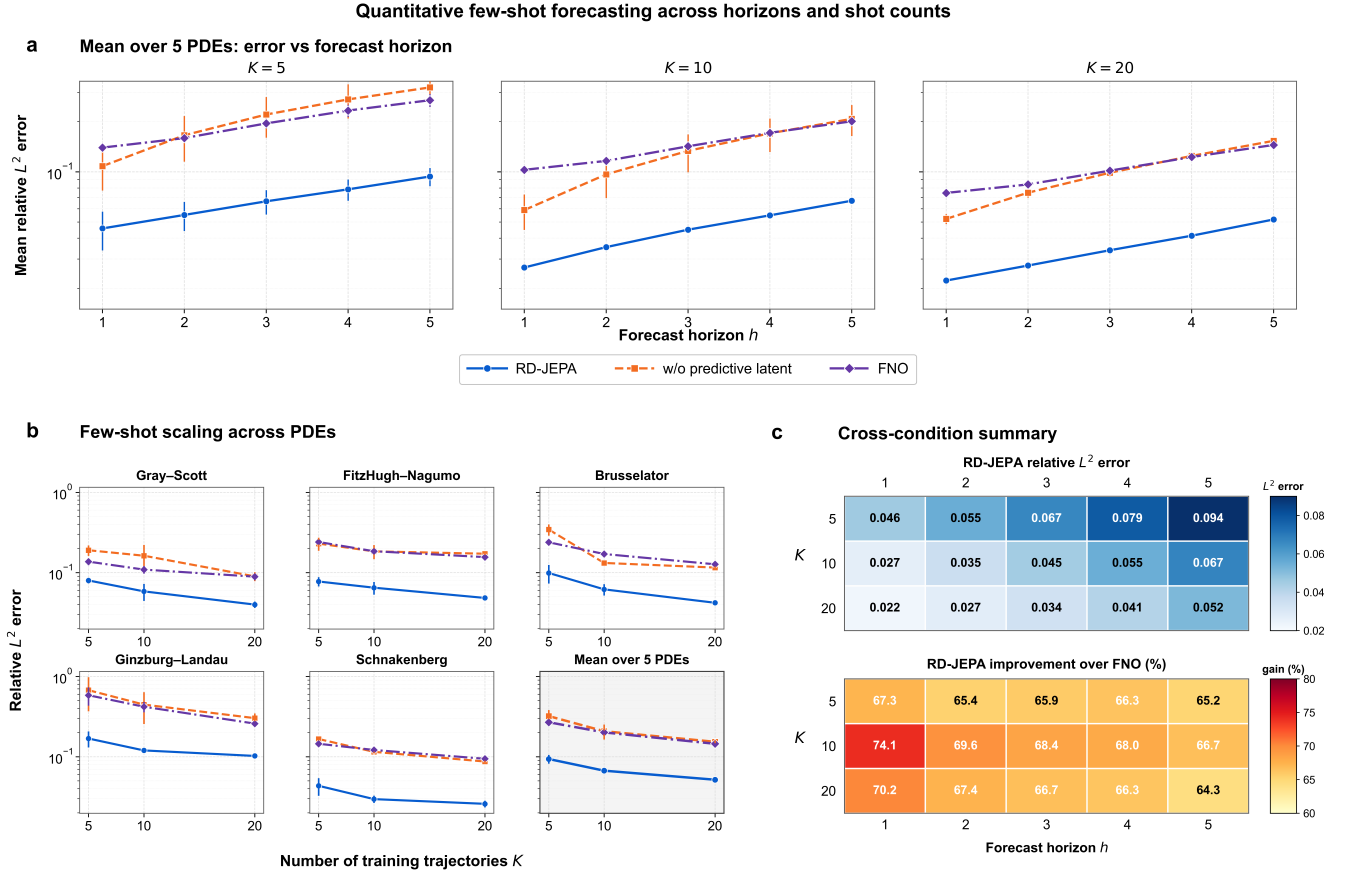}
   \end{minipage}
\caption{
\textbf{Dependence of source-system forecasting accuracy on adaptation budget and forecast horizon.}
{\textbf{a}, Mean relative $L^2$ error as a function of forecast horizon $h\in\{1,\ldots,5\}$ for adaptation budgets $K=5$, $10$ and $20$. RD-JEPA is compared with the no-predictive-latent control (orange) and an equation-specific FNO (purple). Each run-level value is an equal-weight mean across the five source equations after averaging over $300$ fixed test trajectories per equation. Points and error bars show the mean and sample standard deviation across three paired support-selection runs. The vertical axis is logarithmic.
\textbf{b}, Equation-level few-trajectory scaling at $h=5$. Relative $L^2$ error is plotted against $K$ for each source equation and for the equal-weight mean across equations. Points and error bars again show the mean and sample standard deviation across the three runs. Both axes are logarithmic.
\textbf{c}, Summary across all adaptation budgets and forecast horizons. The upper heatmap reports the across-run mean relative $L^2$ error of RD-JEPA. The lower heatmap reports its percentage reduction relative to FNO, $100\left(1-\frac{\mathrm{err}_{\mathrm{RD\mbox{-}JEPA}}}{\mathrm{err}_{\mathrm{FNO}}}\right)$. Positive values indicate lower error for RD-JEPA. Equation- and split-resolved across-run summaries are reported in Supplementary Tables~\ref{tab:supp_source_rel_k5}--\ref{tab:supp_source_split_summary}, with individual run-level values in Supplementary Data~1.}
}
\label{fig:source_scaling}
\end{figure}

Across all adaptation budgets \(K\in\{5,10,20\}\) and forecast horizons \(h\in\{1,\ldots,5\}\), the mean forecasting error increased with the horizon for all three models, while the RD-JEPA mean decreased as additional adaptation trajectories were provided (Fig.~\ref{fig:source_scaling}a,b). RD-JEPA had the smallest mean relative discrete \(\ell^2\) error across the five source systems in every evaluated budget–horizon configuration. Its paired percentage reduction relative to FNO was positive throughout the corresponding \(K\)-by-\(h\) grid (Fig.~\ref{fig:source_scaling}c). These results show that the observed source-system advantage was not limited to a single adaptation budget or forecast horizon.

Complete equation-, adaptation-budget-, and forecast-horizon-resolved results are reported in Supplementary Note 4. Relative discrete \(\ell^2\) field errors are given in Supplementary Tables S6–S8, mean absolute spatial first-difference errors in Supplementary Tables S9–S11, and split-resolved results in Supplementary Table S12. The three run-level values underlying these summaries are provided in Supplementary Data 1.

The representative forecasts in Fig.~\ref{fig:source_qualitative} illustrate how the aggregate error measurements appear in the predicted fields. For FitzHugh–Nagumo, the largest pointwise errors occur near rapidly evolving interfaces. The additional Gray–Scott, complex Ginzburg–Landau, Schnakenberg, and Brusselator examples show the corresponding spatial structures at \(K=5\) and \(h=5\). These examples are illustrative; the quantitative conclusions are based on all 300 test trajectories for each equation.

\subsection*{Predictive-latent representations transfer to unseen reaction laws}
\label{sec:results_ood}

We next tested whether reuse survived a change in the governing reaction law rather than only variation within a source equation. Starting from the same source-pretrained checkpoint, we adapted RD-JEPA to Lambda--Omega, Barkley and Oregonator using $K\in\{1,5,10\}$ target-system trajectories. The reaction terms and trajectories of all three equations were excluded from pretraining. For each equation and value of $K$, three paired runs shared prespecified support sets across models and used the same 300 test trajectories.

\begin{figure}[!htb]
\centering
    \begin{minipage}{\textwidth}
     \centering
     \includegraphics[width=\linewidth]{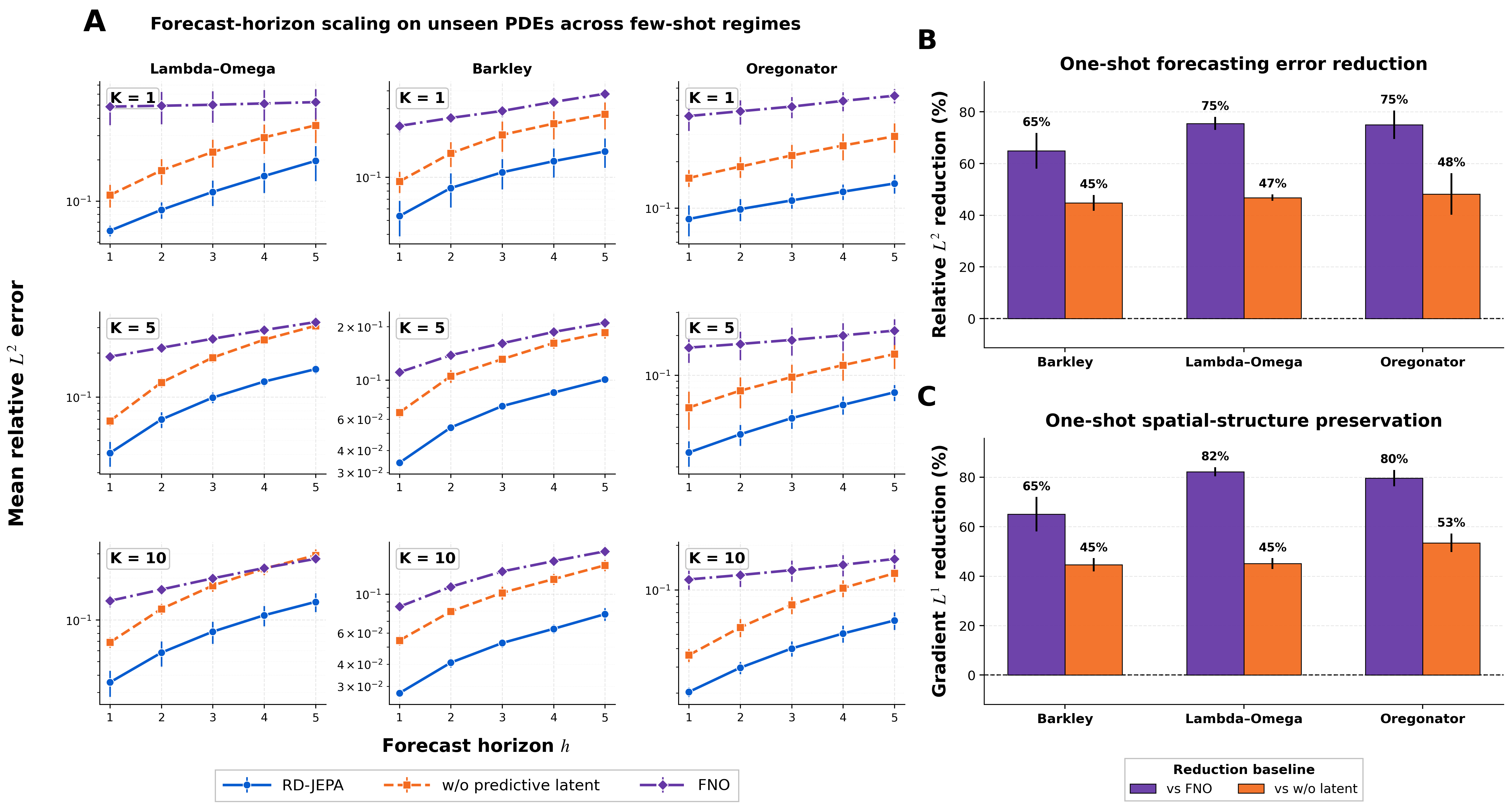}
   \end{minipage}
\caption{
\textbf{Few-trajectory forecasting on governing equations excluded from pretraining.}
{\textbf{a}, Mean relative $L^2$ error versus forecast horizon $h=1,\ldots,5$ for Lambda--Omega, Barkley and Oregonator (columns) after adaptation with $K=1$, $5$ and $10$ trajectories (rows). RD-JEPA (blue) is compared with the no-predictive-latent control (orange) and FNO (purple). Each run-level value is averaged over the same $300$ held-out test trajectories. Points and error bars show the mean and sample standard deviation across three paired support-selection runs; vertical axes are logarithmic.
\textbf{b}, Percentage reduction in horizon-averaged relative $L^2$ error at $K=1$, computed within each matched run as $100\left(1-\frac{\mathrm{err}_{\mathrm{RD\mbox{-}JEPA}}}{\mathrm{err}_{\mathrm{baseline}}}\right)$, relative to FNO (purple) and the control (orange). Bars and error bars show the mean and sample standard deviation across the three paired run-level reductions.
\textbf{c}, Corresponding paired reduction in horizon-averaged gradient $L^1$ error. Positive values indicate lower error for RD-JEPA. The horizon-averaged full-model--control comparison is reported in Supplementary Table~\ref{tab:supp_heldout_predictive_latent_control}, and the horizon-resolved run-level values underlying the figure are provided in Supplementary Data~2.}
}
\label{fig:ood_quantitative}
\end{figure}

RD-JEPA had the lowest mean relative $L^2$ error among RD-JEPA, FNO and the no-predictive-latent control for every evaluated target equation, adaptation budget and forecast horizon (Fig.~\ref{fig:ood_quantitative}a). This ordering was already present after adaptation with one trajectory and persisted through $h=5$. At $K=1$, horizon-averaged relative $L^2$ and gradient $L^1$ errors were lower for RD-JEPA than for both comparators (Fig.~\ref{fig:ood_quantitative}b,c). Because none of the target reaction laws or trajectories contributed to pretraining, these results support equation-level transfer of the learned representation rather than reuse confined to source-equation parameter regimes.

The independently trained no-predictive-latent control removes the trajectory-specific encoder--predictor pathway while retaining the physical-context decoder route and lead-time conditioning. RD-JEPA retained lower mean field and gradient errors under this intervention, supporting the contribution of the complete predictive-latent representation pathway to few-trajectory transfer (Supplementary Note~\ref{supp:note_heldout_results} and Supplementary Table~\ref{tab:supp_heldout_predictive_latent_control}).

\begin{figure}[!htb]
\centering
    \begin{minipage}{\textwidth}
     \centering
     \includegraphics[width=\linewidth]{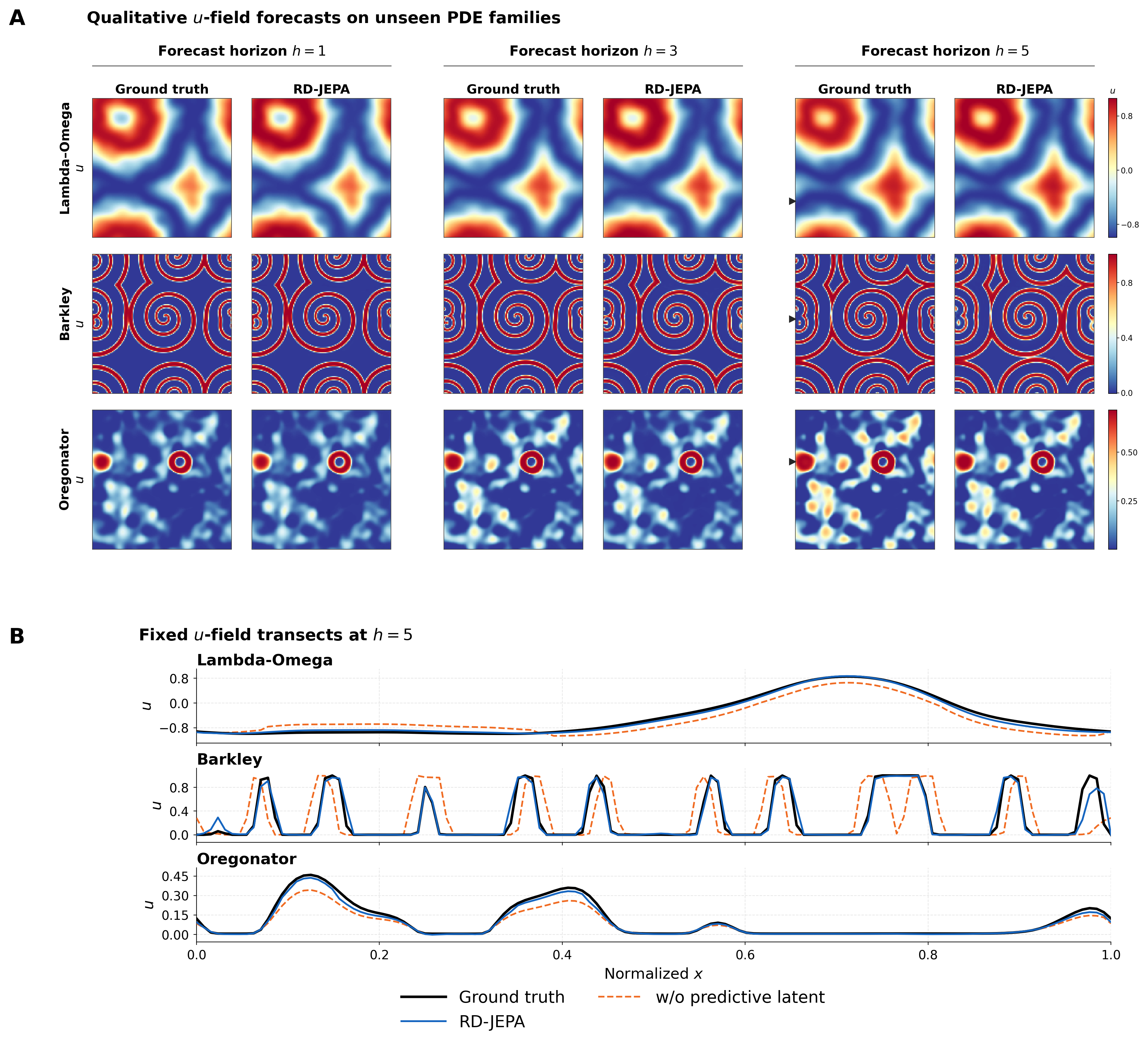}
   \end{minipage}
\caption{
\textbf{Representative one-trajectory forecasts on governing equations excluded from pretraining.}
{\textbf{a}, Ground-truth and RD-JEPA forecasts of the $u$ field for Lambda--Omega, Barkley and Oregonator after adaptation with one target trajectory, shown at forecast horizons $h=1$, $3$ and $5$. Ground truth and prediction share a common colour scale within each equation. The dashed line in each $h=5$ ground-truth field marks the spatial transect used in panel~b.
\textbf{b}, Values of the $u$ field along the indicated transects at $h=5$, plotted against normalized spatial coordinate. Black solid lines denote ground truth, blue solid lines RD-JEPA and orange dashed lines the no-predictive-latent control. Aggregate three-run results over all $300$ held-out test trajectories are reported in Fig.~\ref{fig:ood_quantitative} and Supplementary Table~\ref{tab:supp_heldout_predictive_latent_control}.}
}
\label{fig:ood_visualization}
\end{figure}

Representative one-trajectory forecasts connect the aggregate errors to the propagation of fronts, waves and interfaces in the three target systems (Fig.~\ref{fig:ood_visualization}). Together with the test-set summaries, these examples show how a representation pretrained on five source systems supports forecasts after adaptation to three previously unseen reaction laws.

\subsection*{Cross-equation transfer persists across model classes and controlled training schemes}
\label{sec:results_broad_baselines}

To determine whether the transfer result depended on FNO as the supervised comparator, we evaluated four additional equation-specific surrogates on the three target equations: LNO \cite{cao2024laplace}, ReViT \cite{wei2026revit}, RieszNO \cite{liu2026rieszno} and CNextU-Net \cite{ohana2024the}. Together with FNO \cite{li2021fourier}, these models span spectral and transform-domain neural operators, an equivariant transformer and a convolutional encoder--decoder. Table~\ref{tab:operator_comparison} reports horizon-averaged relative $L^2$ and gradient $L^1$ errors for $K\in\{1,5,10\}$ across the three paired support-set selections. RD-JEPA alone reused a representation pretrained on the five source equations.

RD-JEPA had the lowest mean on both metrics in all nine equation--budget settings (Table~\ref{tab:operator_comparison}). For Lambda--Omega at $K=10$, its mean relative $L^2$ error was essentially indistinguishable from that of RieszNO, supporting comparable performance in this setting. In the other eight settings, RD-JEPA had a lower mean relative $L^2$ error than the strongest external supervised baseline. Persistence across spectral, transform-domain, equivariant and convolutional surrogates makes the transfer result unlikely to be an artefact of selecting FNO as the comparator. The three run-level values underlying each entry are provided in Supplementary Data~3.

An architecture-matched control tested whether architecture alone could reproduce the transfer result when the encoder--predictor--decoder was trained end to end from random initialization. RD-JEPA had lower mean relative $L^2$ and gradient $L^1$ errors in all nine target-equation--budget settings (Supplementary Note~\ref{supp:note_robustness} and Supplementary Table~\ref{tab:supp_scratch_control}). Under matched support sets, downstream objectives and update budgets, source pretraining therefore provided transferable information that the same architecture did not recover from the limited target data alone.

We finally tested adaptation beyond the periodic boundary conditions used during pretraining. For Barkley dynamics with homogeneous Neumann, Robin and Dirichlet conditions, RD-JEPA had the lowest mean relative $L^2$ and gradient $L^1$ errors at $K=1$, $5$ and $10$ (Supplementary Table~\ref{tab:supp_boundary_shift}). The result extends the observed transfer to a combined shift in boundary conditions and diffusion discretization within Barkley at the evaluated domain, resolution and simulation protocol.

\begin{table}[ht]
\centering
\small
\setlength{\tabcolsep}{4.5pt}
\caption{
{\textbf{Forecasting accuracy on governing equations excluded from pretraining.} $K$ denotes the number of independent target-system trajectories used for adaptation. Each run-level score is obtained by first averaging each trajectory's error over $h=1,\ldots,5$ and then averaging across the same $300$ held-out test trajectories. For every model, values are the mean $\pm$ sample standard deviation across three paired support-selection runs. Values are multiplied by $100$, and lower values are better. Boldface marks the lowest numerical mean for each equation and adaptation budget; no formal hypothesis tests were performed. The paired horizon-averaged comparison between RD-JEPA and the no-predictive-latent control is reported in Supplementary Table~\ref{tab:supp_heldout_predictive_latent_control}.}
}
\label{tab:operator_comparison}

\begin{tabular}{llccc@{\hspace{8pt}}ccc}
\toprule
& &
\multicolumn{3}{c}{\textbf{Relative $L^2$ $\downarrow$ $(\times10^{-2})$}} &
\multicolumn{3}{c}{\textbf{Gradient $L^1$ $\downarrow$ $(\times10^{-2})$}} \\
\cmidrule(lr){3-5}
\cmidrule(lr){6-8}
\textbf{PDE} & \textbf{Model} &
\textbf{$K=1$} & \textbf{$K=5$} & \textbf{$K=10$} &
\textbf{$K=1$} & \textbf{$K=5$} & \textbf{$K=10$} \\
\midrule

Lambda--Omega 
& RD-JEPA 
& $\mathbf{12.24 \pm 2.58}$ 
& $\mathbf{9.85 \pm 0.79}$ 
& $\mathbf{8.38 \pm 1.46}$
& $\mathbf{0.68 \pm 0.12}$ 
& $\mathbf{0.52 \pm 0.05}$ 
& $\mathbf{0.45 \pm 0.08}$ \\

& FNO 
& $50.53 \pm 13.06$ 
& $25.48 \pm 1.09$ 
& $20.29 \pm 1.59$
& $3.87 \pm 1.02$ 
& $1.89 \pm 0.07$ 
& $1.50 \pm 0.08$ \\

& LNO 
& $34.56 \pm 8.15$ 
& $21.07 \pm 1.26$ 
& $16.65 \pm 2.16$
& $2.51 \pm 0.55$ 
& $1.57 \pm 0.11$ 
& $1.37 \pm 0.16$ \\

& ReViT 
& $22.64 \pm 4.11$ 
& $15.35 \pm 0.72$ 
& $13.61 \pm 1.37$
& $2.85 \pm 0.22$ 
& $3.24 \pm 0.14$ 
& $3.04 \pm 0.13$ \\

& RieszNO 
& $19.20 \pm 4.72$ 
& $13.41 \pm 0.88$ 
& $8.47 \pm 1.14$
& $1.26 \pm 0.28$ 
& $0.88 \pm 0.06$ 
& $0.47 \pm 0.06$ \\

& CNextU-Net 
& $25.52 \pm 6.03$ 
& $21.94 \pm 1.32$ 
& $16.54 \pm 2.16$
& $1.72 \pm 0.38$ 
& $2.02 \pm 0.14$ 
& $1.92 \pm 0.22$ \\

\midrule

Barkley 
& RD-JEPA 
& $\mathbf{10.53 \pm 2.55}$ 
& $\mathbf{6.90 \pm 0.24}$ 
& $\mathbf{5.23 \pm 0.32}$
& $\mathbf{1.09 \pm 0.27}$ 
& $\mathbf{0.74 \pm 0.04}$ 
& $\mathbf{0.58 \pm 0.04}$ \\

& FNO 
& $29.86 \pm 2.52$ 
& $16.13 \pm 0.75$ 
& $13.15 \pm 0.33$
& $3.11 \pm 0.26$ 
& $1.66 \pm 0.07$ 
& $1.37 \pm 0.04$ \\

& LNO 
& $34.58 \pm 3.15$ 
& $17.65 \pm 0.78$ 
& $16.39 \pm 0.62$
& $1.97 \pm 0.28$ 
& $1.59 \pm 0.07$ 
& $1.59 \pm 0.08$ \\

& ReViT 
& $22.03 \pm 1.03$ 
& $14.72 \pm 0.35$ 
& $11.08 \pm 0.28$
& $2.24 \pm 0.12$ 
& $1.50 \pm 0.03$ 
& $1.15 \pm 0.03$ \\

& RieszNO 
& $46.75 \pm 4.07$ 
& $24.59 \pm 1.05$ 
& $21.99 \pm 0.69$
& $4.09 \pm 0.38$ 
& $3.09 \pm 0.13$ 
& $3.00 \pm 0.10$ \\

& CNextU-Net 
& $30.85 \pm 3.10$ 
& $17.40 \pm 0.73$ 
& $13.89 \pm 0.56$
& $3.97 \pm 0.42$ 
& $2.16 \pm 0.10$ 
& $1.87 \pm 0.07$ \\
\midrule

Oregonator 
& RD-JEPA 
& $\mathbf{11.36 \pm 1.34}$ 
& $\mathbf{4.83 \pm 0.79}$ 
& $\mathbf{4.05 \pm 0.47}$
& $\mathbf{0.47 \pm 0.05}$ 
& $\mathbf{0.26 \pm 0.06}$ 
& $\mathbf{0.22 \pm 0.02}$ \\

& FNO 
& $46.17 \pm 7.07$ 
& $18.79 \pm 4.42$ 
& $13.76 \pm 2.21$
& $2.32 \pm 0.27$ 
& $1.04 \pm 0.15$ 
& $0.73 \pm 0.08$ \\

& LNO 
& $29.41 \pm 3.42$ 
& $20.66 \pm 3.87$ 
& $13.39 \pm 1.65$
& $1.29 \pm 0.13$ 
& $0.93 \pm 0.14$ 
& $0.69 \pm 0.07$ \\

& ReViT 
& $26.73 \pm 1.96$ 
& $20.49 \pm 1.10$ 
& $20.03 \pm 1.38$
& $2.64 \pm 0.07$ 
& $2.01 \pm 0.04$ 
& $2.01 \pm 0.03$ \\

& RieszNO 
& $27.42 \pm 3.28$ 
& $19.14 \pm 2.91$ 
& $16.50 \pm 1.90$
& $1.35 \pm 0.16$ 
& $1.00 \pm 0.17$ 
& $0.86 \pm 0.09$ \\

& CNextU-Net 
& $26.01 \pm 2.83$ 
& $16.76 \pm 3.31$ 
& $13.83 \pm 1.68$
& $2.08 \pm 0.20$ 
& $1.11 \pm 0.19$ 
& $0.90 \pm 0.11$ \\

\bottomrule
\end{tabular}
\end{table}

\section*{Discussion}
Our results show that predictive-latent pretraining can transform trajectories from several reaction--diffusion equations into a reusable dynamical representation that remains effective after adaptation to previously unseen reaction laws. Reuse on the source equations first established data efficiency, whereas transfer to Lambda--Omega, Barkley and Oregonator showed that the representation was not confined to source-equation parameter regimes. The mean-error ordering persisted across adaptation budgets, forecast horizons and specialized surrogate architectures, indicating that the transferable signal was not specific to one downstream comparator.

This result extends cross-system PDE pretraining beyond objectives that predict or reconstruct physical fields directly. Multiple Physics Pretraining, Poseidon and masked function-space pretraining learn reusable models or representations through field-level prediction or reconstruction \cite{mccabe2024multiple,herde2024poseidon,rahman2024pretraining}, whereas RD-JEPA predicts future representations from trajectory context. By moving the predictive target from field space to representation space, RD-JEPA tests a different locus for shared dynamical information, following the JEPA principle developed for visual representation learning \cite{assran2023ijepa,bardes2024vjepa}. Physical JEPA studies have also considered parameter inference and physics-informed surrogate pretraining \cite{qu2026representation,yee2026pijepa}. The present work identifies representation-space prediction as a viable pretraining target for shared learning across reaction laws followed by few-trajectory full-field forecasting. Direct comparisons among predictive, reconstructive and contrastive objectives offer a natural route for characterizing their relative transfer properties.

RD-JEPA also embeds a problem-specific hypothesis through complementary diffusion-like and reaction-like latent pathways. This decomposition serves as an inductive bias rather than an exact operator splitting and requires no equation identifier, coefficient or symbolic governing equation. The no-predictive-latent and architecture-matched scratch controls evaluate the complete source-pretrained pathway, whose transfer benefit persists under both comparisons. Component-wise analyses could further resolve how the two predictor pathways and the pretraining objective interact.

The present scope covers deterministic, two-component reaction--diffusion systems at one two-dimensional resolution and five directly predicted horizons. The non-periodic Barkley experiment evaluates a coupled shift in boundary conditions and diffusion discretization. RD-JEPA operates through few-trajectory target adaptation rather than zero-shot inference. Variability estimates capture support-set selection for one predictive-pretraining checkpoint and one fixed model-training random-number-generator state. Longer rollouts, additional resolutions and repeated pretraining runs will be needed to characterize stability and variability beyond this setting.

Within these limits, future-state prediction in representation space provides a viable route to domain-specific reusable dynamics models. A simulation corpus from several related equations can be amortized into one representation and repeatedly specialized when target-system trajectories are scarce. In this operational sense, RD-JEPA constitutes a predictive latent world model and a step towards foundation modelling for reaction--diffusion dynamics.

\section*{Methods}

\subsection*{Reaction--diffusion datasets and experimental design}

We considered two-field reaction--diffusion systems of the form
\begin{equation}
\partial_t\mathbf{u}=\mathbf{D}\nabla^2\mathbf{u}
+\mathbf{R}(\mathbf{u};\boldsymbol{\theta}),
\qquad \mathbf{u}=(u,v)^{\mathsf T}.
\label{eq:methods_rd_general}
\end{equation}
{Predictive pretraining used Gray--Scott, FitzHugh--Nagumo, Brusselator, complex Ginzburg--Landau and Schnakenberg dynamics. Lambda--Omega, Barkley and Oregonator were held out at the governing-equation level: their reaction terms and trajectories were excluded from pretraining. Each primary-benchmark trajectory contained 40 two-channel fields on a final $128\times128$ periodic grid. The governing equations, coefficient and initial-condition distributions, numerical solvers and temporal sampling intervals are provided in Supplementary Note~\ref{supp:note_dataset} and Supplementary Table~\ref{tab:supp_numerics}.}

{For each source equation, we generated 1,000 source-pool trajectories used for predictive pretraining and downstream support selection, 50 validation trajectories and 300 test trajectories. The test set comprised 100 in-distribution trajectories, 100 trajectories from held-out coefficient regimes and 100 trajectories from held-out initial-condition families. Each equation excluded from pretraining used an independently simulated pool of 1,000 trajectories for downstream adaptation and a fixed set of 300 additional test trajectories. All partitions were trajectory-disjoint, and every temporal window inherited the split of its parent trajectory.}

{Source-equation adaptation used $K\in\{5,10,20\}$ trajectories, whereas equation-level transfer used $K\in\{1,5,10\}$. Here, $K$ counts complete trajectories rather than temporal windows. We evaluated three prespecified support-set selections. Within each selection, support sets were nested across $K$ and shared by all methods, and each adapted or from-scratch model was evaluated on the same 300 test trajectories. The predictive-pretraining checkpoint and model-training random-number-generator state were fixed across selections, so only the selected support trajectories varied. A fixed channel-wise affine transformation, estimated without test data, was applied consistently across methods. Support-selection seeds, the fixed training seeds and preprocessing are reported in Supplementary Note~\ref{supp:note_experiments}.}

{A separate boundary-condition stress test used Barkley dynamics with homogeneous Neumann, Robin or Dirichlet conditions. Each boundary-specific dataset contained 20 adaptation trajectories, five validation trajectories and 300 test trajectories; $K\in\{1,5,10\}$ trajectories were selected for adaptation. These trajectories did not enter predictive pretraining. Common index-level randomization aligned coefficient and initial-condition draws before boundary-specific quality screening, so the three datasets represent matched generating distributions rather than exactly paired trajectories. Within each boundary condition, all methods received identical support trajectories and were evaluated on the same test trajectories. Dataset construction is detailed in Supplementary Note~\ref{supp:note_dataset}.}

\subsection*{Joint-embedding predictive pretraining}

{RD-JEPA follows the joint-embedding predictive formulation used for image and video representation learning \cite{assran2023ijepa,bardes2024vjepa}. Given four consecutive states $\mathbf{X}^{c}_{t}=(\mathbf{x}_{t-3},\ldots,\mathbf{x}_{t})$, an online encoder $E_{\theta}$ produced 256 spatial context tokens of width 512. An exponential-moving-average target encoder $E_{\xi}$ represented the target field $\mathbf{x}_{t+h}$, and a predictor $P_{\phi}$ estimated its target-patch representations from the context, patch position and forecast horizon $h$, measured in stored-frame intervals. The target branch received no gradients and followed the online encoder through an exponential moving average. One model was shared across the five source equations and received no equation identity, coefficient vector or symbolic governing equation.}

Each channel was embedded with non-overlapping $8\times8$ patches, yielding a fused $16\times16$ token field. Temporal cross-attention aggregated the four context frames, after which a multiscale U-shaped operator encoder combined global, spectral and local spatial mixing. The latent predictor alternated target self-attention and context cross-attention with two reaction--diffusion-aligned updates. For context token $\mathbf{z}_{c,i}$, the diffusion-like feature was
\begin{equation}
\Delta_{\mathrm{lat}}\mathbf{z}_{c,i}
=\frac{1}{4}\sum_{j\in\mathcal{N}(i)}\mathbf{z}_{c,j}
-\mathbf{z}_{c,i},
\label{eq:methods_latent_laplacian}
\end{equation}
{where $\mathcal{N}(i)$ contains the four periodic nearest neighbours in the primary benchmark. A pointwise pathway supplied a reaction-like update, and both pathways were conditioned on a continuous embedding of $h$. For the non-periodic Barkley stress test, the predictor retained this pretrained periodic neighbour map and received no boundary-condition label or explicit target-domain boundary operator. Complete block dimensions, positional encodings, conditioning operations and parameter counts are given in Supplementary Note~\ref{supp:note_model} and Supplementary Table~\ref{tab:supp_jepa_architecture}.}

Training batches were balanced across the five source equations. Pretraining sampled future targets up to eight stored-frame intervals ahead and combined future-block, future-tube and same-time masks. The online encoder and predictor minimized the representation-space objective
\begin{equation}
\mathcal{L}_{\mathrm{JEPA}}
=\frac{1}{BMD}
\sum_{b=1}^{B}\sum_{m=1}^{M}
\left\|
\widetilde{\widehat{\mathbf{z}}}_{b,m}
-\operatorname{sg}\!\left(
\widetilde{\mathbf{z}}_{b,m}
\right)
\right\|_2^2,
\label{eq:methods_jepa_loss}
\end{equation}
where $B$ is the batch size, $M=256$, $D=512$, $\operatorname{sg}$ denotes stop-gradient and the tildes denote the feature transformation defined in Supplementary Note~\ref{supp:note_model}. No field decoder was attached during pretraining. We trained for 200,000 optimization steps; the masking distribution and remaining optimization settings are listed in Supplementary Tables~\ref{tab:supp_target_sampling} and~\ref{tab:supp_optimization}.

\subsection*{Few-trajectory adaptation and comparison models}

For downstream forecasting, the online encoder was frozen and the target encoder was discarded. The pretrained predictor was fine-tuned jointly with a newly initialized dense decoder. The predictor was queried independently at $h\in\{1,\ldots,5\}$, and the decoder fused its $16\times16$ latent field with multiscale features from the latest context state. The model therefore predicted each requested horizon directly rather than through autoregressive rollout. One predictor--decoder pair was adapted for each equation, value of $K$ and support-set selection.

RD-JEPA and the no-predictive-latent control used a supervised objective combining pointwise, relative-field, spatial-gradient and Fourier-magnitude errors, with uniform weighting over the five horizons. Each primary-benchmark adaptation used 5,000 optimization steps. All comparison models and controls used the same supervised objective and loss weights as RD-JEPA. Optimizer settings, loss weights and the complete decoder specification are reported in Supplementary Notes~\ref{supp:note_model} and~\ref{supp:note_experiments} and Supplementary Tables~\ref{tab:supp_jepa_architecture}--\ref{tab:supp_baseline_optimization}.

The no-predictive-latent control was trained independently. It removed the JEPA encoder--predictor pathway, retained the same last-frame decoder pathway and supplied only a spatially constant lead-time embedding to the latent interface. The control therefore contained no trajectory-specific predictive latent and evaluates the contribution of the complete sample-dependent predictive-latent pathway under the downstream forecasting protocol. The supervised comparisons additionally included FNO, LNO, RieszNO, ReViT and CNextU-Net. Each baseline received the same support trajectories, four-frame context and five direct target horizons as RD-JEPA.

The architecture-matched scratch control retained the complete encoder--predictor--decoder architecture but trained it end to end from random initialization on the three equations excluded from pretraining. It used the same support trajectories, direct horizons, downstream objective, 5,000-update budget and fixed model-training random-number-generator state as RD-JEPA, but no pretrained tensors, exponential-moving-average target branch or JEPA objective. The final-update checkpoint was evaluated without validation- or test-based checkpoint selection. This architecture-matched comparison evaluates source-pretrained adaptation against end-to-end training from random initialization under identical support sets, forecast horizons, downstream objectives and update budgets.

For the non-periodic Barkley stress test, all models used 5,000 optimization steps. Method-specific objectives, optimization budgets and architecture details are given in Supplementary Notes~\ref{supp:note_experiments} and~\ref{supp:note_robustness} and Supplementary Table~\ref{tab:supp_baseline_optimization}.

\subsection*{Evaluation and statistical aggregation}

All errors were computed on the original field scale. The principal metrics were relative field error and first-order spatial-gradient error,
\begin{equation}
E_{\mathrm{rel}L^2}
=\frac{\|\widehat{\mathbf{x}}-\mathbf{x}\|_2}
{\|\mathbf{x}\|_2+10^{-8}},
\qquad
E_{\nabla L^1}
=\operatorname{MAE}(\delta_x\widehat{\mathbf{x}},\delta_x\mathbf{x})
+\operatorname{MAE}(\delta_y\widehat{\mathbf{x}},\delta_y\mathbf{x}).
\label{eq:methods_metrics}
\end{equation}
Here, $\delta_x$ and $\delta_y$ denote adjacent first differences along the two spatial axes, without a wrap-around term. The mean absolute errors are averaged over both field channels and all valid neighbouring grid pairs. The same metric definition is used for the periodic and non-periodic datasets.
Mean-squared error was retained as a secondary diagnostic for the primary source- and target-equation benchmarks and is supplied in Supplementary Data~1--3.

Every valid temporal window was evaluated. With 40 frames, a four-frame context and a maximum horizon of five, each trajectory contributed 32 windows. Errors were first averaged over windows within each trajectory and then with equal weight over the 300 test trajectories. Horizon-averaged scores assigned equal weight to $h=1,\ldots,5$. Each displayed value was computed separately for each support-set selection and then summarized as the mean and sample standard deviation across the three run-level means. Thus, $n=3$ denotes support-set selections, not test trajectories or independent pretraining runs. The same 300 test trajectories were reused across selections, and the predictive-pretraining checkpoint and model-training random-number-generator state were fixed. The reported standard deviation therefore quantifies sensitivity to support-set selection. No formal hypothesis tests were performed. Complete aggregation definitions and source-data mappings are provided in Supplementary Note~\ref{supp:note_experiments}.
\section*{Acknowledgments}

This work was partially supported by the National Natural Science Foundation of China (72495131, 82441027), Guangdong Provincial Key Laboratory of Mathematical Foundations for Artificial Intelligence (2023B1212010001), Shenzhen Stability Science Program, and the Shenzhen Science and Technology Program under grant no. JCYJ20250604141043020.

\section*{Author Contributions Statement}
C.S. and M.Y. conceived the study. C.S. developed the method, implemented the models, generated the datasets, performed the experiments and prepared the figures. C.S. and M.Y. analysed the results and wrote the manuscript. M.Y. supervised the project.
\section*{Data availability}
{The source data underlying the quantitative figures and tables are organized as Supplementary Data~1--5 and will be deposited in a public archival repository. For each quantitative display, the deposit will contain the run-level values and their reported across-run mean and sample standard deviation. Run-level aggregation follows the averaging unit stated in the corresponding caption: 300 trajectories for pooled equation results, 100 trajectories for each split-resolved source result and 300 boundary-specific trajectories for the Barkley stress test. The deposit will also include the numerical arrays underlying the qualitative fields and transects, prespecified dataset splits, support-set indices and plotting inputs. Simulation and evaluation code, parameter files and recorded random seeds will permit regeneration of the trajectory-level benchmark. The persistent repository identifier will be added before publication.}

\section*{Code availability}
{Code for simulation, predictive pretraining, downstream adaptation, baseline training, evaluation and figure generation will be made available to editors and reviewers through an anonymous repository and archived publicly upon publication. The release will include environment specifications, configuration files and instructions linking each reported result to its source-data file. A persistent version identifier will be added before publication.}

\section*{Competing interests}

The authors declare no competing interests.

\bibliography{reference}

\clearpage
\renewcommand{\thetable}{S\arabic{table}}
\renewcommand{\thefigure}{S\arabic{figure}}
\renewcommand{\theequation}{S\arabic{equation}}
\renewcommand{\theHtable}{supp.table.\arabic{table}}
\renewcommand{\theHfigure}{supp.figure.\arabic{figure}}
\renewcommand{\theHequation}{supp.equation.\arabic{equation}}
\setcounter{table}{0}
\setcounter{figure}{0}
\setcounter{equation}{0}
\setcounter{supplementarynote}{0}
\renewcommand{\arraystretch}{1.12}

\thispagestyle{empty}
\begingroup
\centering
\vspace*{1.5cm}

{\LARGE Supplementary Information for\par}
\vspace{1.2cm}

{\Large \textbf{{RD-JEPA: Predictive latent pretraining for few-trajectory transfer across reaction--diffusion equations}}\par}
\vspace{1.2cm}

{\large Chenhao Si$^{1}$, Ming Yan$^{1,*}$\par}
\vspace{0.8cm}

{\normalsize $^{1}$School of Data Science, The Chinese University of Hong Kong, Shenzhen, Shenzhen, China\par}
\vspace{1.5cm}

{\normalsize $^{*}$Corresponding author: \texttt{yanming@cuhk.edu.cn}\par}
\vfill
\par
\endgroup

\clearpage
\tableofcontents
\clearpage

\suppnote{Supplementary Note 1: Reaction--diffusion systems and dataset construction}
\label{supp:note_dataset}

\suppsubsection{S1.1 Benchmark composition, evaluation regimes, and notation}
\label{supp:dataset_overview}
We constructed a benchmark comprising eight two-dimensional, two-component reaction–diffusion systems. Trajectories from Gray–Scott, FitzHugh–Nagumo, Brusselator, complex Ginzburg–Landau, and Schnakenberg were used for predictive pretraining. Lambda–Omega, Barkley, and Oregonator were reserved for downstream transfer; neither their nonlinear reaction operators nor any trajectories generated by them were included in pretraining.

All systems are written in the form
\begin{equation}
\partial_t\mathbf{u}(\mathbf{x},t)=\mathbf{D}\nabla^2\mathbf{u}(\mathbf{x},t)+\mathbf{R}\!\left(\mathbf{u}(\mathbf{x},t);\boldsymbol{\theta}\right),
\qquad \mathbf{u}=(u,v)^{\mathsf T},
\label{eq:supp_general_rd}
\end{equation}
where $\mathbf{D}=\mbox{diag}(D_u,D_v)$ is the diffusion matrix and $\mathbf{R}$ is the nonlinear reaction operator with parameters \(\boldsymbol{\theta}\). Each trajectory in the primary benchmark consists of 40 stored two-channel fields represented on a \(128\times128\) periodic grid and is stored as $\mathbf{X}^{(i)}\in\mathbb{R}^{40\times128\times128\times2}$. The coefficient and initial-condition distributions were chosen to generate a range of spatial and temporal behaviors, including spots, fronts, oscillations, phase defects, and spiral waves.

For each source system, evaluation was performed under three test regimes: in-distribution, coefficient-out-of-distribution, and initial-condition-out-of-distribution. These source-system evaluations are distinct from the held-out-system transfer experiment, in which the complete Lambda–Omega, Barkley, and Oregonator systems were not included in pretraining.

For each source system, we generated a pool of 1,000 trajectories for predictive pretraining and downstream support-set selection, together with 50 validation trajectories and 300 test trajectories. The test set contained 100 trajectories from each of the three evaluation regimes. The 1,000-trajectory source pool, validation set, and test set were mutually disjoint at the trajectory level. The support trajectories used for source-system adaptation were selected from the source pool used during predictive pretraining. For each held-out system, a separate pool of 1,000 trajectories was used for support-set selection, and 300 additional trajectories formed the fixed test set. Every temporal window inherited its parent trajectory's partition.

The forecast horizon \(h\) is measured in stored-frame intervals. Because the internal integration step and frame-saving interval are system-dependent, the same value of \(h\) does not necessarily correspond to the same physical elapsed time across systems.

\suppsubsection{S1.2 Source systems used for predictive pretraining}
\label{supp:source_equations}
Unless otherwise stated, each scalar model parameter specified by an interval below was sampled independently from the uniform distribution on that interval.

\paragraph{Gray--Scott.} We considered the Gray–Scott system
\begin{align}
\partial_{\tau}u &= D_u\nabla^2u-uv^2+f(1-u),\\
\partial_{\tau}v &= D_v\nabla^2v+uv^2-(f+k)v.
\label{eq:supp_gray_scott}
\end{align}
This system supports spot, stripe, maze, and moving-pattern regimes~\cite{pearson1993complex}. The implementation used the rescaled time variable \(t=\tau/1000\), with all coefficients transformed consistently. We first selected a parameter center $(f_0,k_0)$ from
\[
(0.008,0.046),\ (0.020,0.056),\ (0.040,0.060),\ (0.029,0.057),\ (0.058,0.065).
\]
Then both parameters were perturbed independently by up to \(6\%\). 
The reference diffusion coefficients \(D_u=2\times10^{-5}\) and \(D_v=10^{-5}\) were perturbed independently by up to \(5\%\). Initial conditions consisted of smooth, low-frequency Fourier fields, localized Gaussian perturbations, and their mixtures. Their phases, centers, spatial scales, and amplitudes were randomized.

\paragraph{FitzHugh--Nagumo.} We generated excitable dynamics using the FitzHugh--Nagumo system \cite{fitzhugh1961impulses,nagumo1962active}
\begin{align}
\partial_tu &= \gamma_u\nabla^2u+u-u^3-v+\alpha,\\
\partial_tv &= \gamma_v\nabla^2v+\beta(u-v),
\label{eq:supp_fhn}
\end{align}
with $\gamma_u=1$, $\gamma_v=100$, $\alpha\sim\mathcal{U}(0.005,0.018)$, and $\beta\sim\mathcal{U}(0.22,0.35)$. The initial $u$- and $v$-fields were generated independently as Gaussian random fields, with their standard deviations sampled from \([0.04,0.06]\). 

\paragraph{Brusselator.} We simulated the Brusselator system\cite{prigogine1968symmetry}
\begin{align}
\partial_tu &= D_u\nabla^2u+\rho\!\left[A-(B+1)u+u^2v\right],\\
\partial_tv &= D_v\nabla^2v+\rho\!\left[Bu-u^2v\right],
\label{eq:supp_brusselator}
\end{align}
with $A\in[0.7,1.1]$, $B\in[2.0,2.6]$, $D_u\in[10^{-4},3\times10^{-4}]$, $D_v\in[5\times10^{-4},1.2\times10^{-3}]$, and $\rho\in[8,12]$. For a given pair $(A,B)$, the spatially homogeneous equilibrium is \((u_\ast,v_\ast)=\left(A,\frac{B}{A}\right).\) Initial conditions were smooth perturbations of the equilibrium. Each perturbation combined a low-pass random field, a weak sinusoidal component, and a localized Gaussian component.

\paragraph{Complex Ginzburg--Landau.} Let $\psi=u+\mathrm{i}v$ be a complex-valued field. We considered the complex Ginzburg--Landau equation\cite{aranson2002world}
\begin{equation}
    \partial_t\psi=\varepsilon\nabla^2\psi+\nu\psi-(\gamma_r+\mathrm{i}\gamma_i)|\psi|^2\psi. 
\label{eq:supp_cgl_complex}
\end{equation}
Equivalently, its real and imaginary components satisfy
\begin{align}
\partial_tu &= \varepsilon\nabla^2u+\nu u-(u^2+v^2)(\gamma_{\mathrm r}u-\gamma_{\mathrm i}v),\\
\partial_tv &= \varepsilon\nabla^2v+\nu v-(u^2+v^2)(\gamma_{\mathrm r}v+\gamma_{\mathrm i}u).
\label{eq:supp_cgl_real}
\end{align}
We fixed $\nu=\gamma_{\mathrm r}=1$. The pair $(\varepsilon,\gamma_i)$ was sampled from one of five parameter regimes with different phase-rotation behavior. Across the five regimes, the parameter values covered \(\varepsilon\in[0.002,0.010],~      \gamma_i\in[0.5,9].\) Initial conditions consisted of randomized vortex seeds, noisy single-vortex states, and phase-disk configurations. The initial-condition-out-of-distribution split used noisier states containing multiple vortices.

\paragraph{Schnakenberg.} We used the Schnakenberg system\cite{schnakenberg1979simple}
\begin{align}
\partial_tu &= D_u\nabla^2u+\rho(a-u+u^2v),\\
\partial_tv &= D_v\nabla^2v+\rho(b-u^2v),
\label{eq:supp_schnakenberg}
\end{align}
with $a\in[0.03,0.06]$, $b\in[0.84,0.88]$, $D_u\in[10^{-4},3\times10^{-4}]$, $D_v\in[5\times10^{-4},1.2\times10^{-3}]$ and $\rho\in[8,12]$. For given $a$ and $b$, the spatially homogeneous equilibrium is
\(
    (u_\ast,v_\ast)
    =
    \left(
        a+b,\,
        \frac{b}{(a+b)^2}
    \right).
\) Initial conditions were smooth perturbations of this equilibrium and used the same low-pass random, sinusoidal, and localized components as those used for the Brusselator.

\suppsubsection{S1.3 Held-out systems for equation-level transfer}
\label{supp:target_equations}

The three systems below were excluded entirely from predictive pretraining and were used only for downstream adaptation and evaluation. 

\paragraph{Lambda--Omega.} Letting $r^2=u^2+v^2$. We considered the Lambda--Omega system~\cite{sherratt1994evolution}
\begin{align}
\partial_tu &= D_u\nabla^2u+\rho\!\left[\alpha(1-r^2)u-(\omega_0-\beta r^2)v\right],\\
\partial_tv &= D_v\nabla^2v+\rho\!\left[(\omega_0-\beta r^2)u+\alpha(1-r^2)v\right].
\label{eq:supp_lambda_omega}
\end{align}
The coefficients were sampled from $\alpha\in[0.75,1.25]$, $\omega_0\in[0.85,1.20]$, $\beta\in[0.20,0.70]$, $\rho\in[0.85,1.20]$ and $D_u,D_v\in[2.5\times10^{-4},7.5\times10^{-4}]$. Initial conditions were drawn from three families: spiral-defect states, target waves, and oblique phase waves. Their centers, orientations, spatial frequencies, and smooth perturbations were randomized.

\paragraph{Barkley.} We considered the Barkley system
\begin{align}
\partial_tu &= D_u\nabla^2u+\frac{1}{\epsilon}u(1-u)\!\left(u-\frac{v+b}{a}\right),\\
\partial_tv &= D_v\nabla^2v+u-v.
\label{eq:supp_barkley}
\end{align}
We used the singular diffusive setting $D_v=0$, so that only the activator field $u$ diffuses \cite{barkley1991model}. The remaining coefficients were sampled from $a\in[0.72,0.78]$, $b\in[0.015,0.028]$, $\epsilon\in[0.018,0.024]$ and $D_u\in[0.90,1.10]$. Initial conditions were broken-wave configurations formed from intersecting excited and refractory fronts, with randomized orientation, displacement, width, and smooth-noise perturbations.

\paragraph{Oregonator.} We considered the reduced two-variable Oregonator system
\begin{align}
\partial_tu &= D_u\nabla^2u+\frac{\rho}{\epsilon}\left[u(1-u)-fv\frac{u-q}{u+q}\right],\\
\partial_tv &= D_v\nabla^2v+\rho(u-v),
\label{eq:supp_oregonator}
\end{align}
which is a reduced activator--inhibitor model of Belousov--Zhabotinsky-type chemistry \cite{field1972oscillations,field1974oregonator}. We sampled $\epsilon\in[0.040,0.080]$, $f\in[1.10,1.55]$, $q\in[0.0015,0.0045]$, $D_u\in[4\times10^{-5},1.6\times10^{-4}]$, $D_v\in[10^{-5},8\times10^{-5}]$ and $\rho\in[0.70,1.40]$. Initial conditions were excitation blobs, ring-shaped target waves, or broken wavefronts superimposed on the positive spatially homogeneous equilibrium, together with weak low-pass noise.

\suppsubsection{S1.4 Numerical integration and trajectory screening}
\label{supp:numerics}

All simulations in the primary benchmark used periodic boundary conditions. For Gray--Scott, Brusselator, complex Ginzburg--Landau, Schnakenberg, Lambda--Omega, and Oregonator, the Laplacian was discretized using second-order periodic central finite differences, and the resulting semi-discrete systems were advanced in time using explicit Euler. FitzHugh--Nagumo used a fourth-order periodic finite-difference approximation of the Laplacian together with classical fourth-order Runge--Kutta integration. Barkley was integrated using Strang splitting: the reaction substep was advanced by the explicit midpoint method, whereas the diffusion substep was integrated exactly in Fourier space. Barkley trajectories were simulated on a $256\times256$ grid and subsequently average-pooled to $128\times128$. The numerical and temporal settings are summarized in Table~\ref{tab:supp_numerics}.

\begin{table}[htbp]
\centering
\small
\caption{\textbf{Numerical discretization and temporal sampling in the primary periodic benchmark.} Each retained trajectory contains 40 saved two-channel fields.}
\label{tab:supp_numerics}
\begin{tabularx}{\linewidth}{@{}p{0.145\linewidth}p{0.19\linewidth}Y p{0.105\linewidth}p{0.16\linewidth}@{}}
\toprule
\textbf{System} & \textbf{Domain and grid} & \textbf{Integrator and spatial discretization} & \textbf{Internal $\Delta t$} & \textbf{Saved window} \\
\midrule
Gray--Scott & $[-1,1)^2$, $128\times128$ & Explicit Euler; second-order periodic central differences & $10^{-4}$ & $t\in[0.05,1.00]$ \\
FitzHugh--Nagumo & Periodic $128\times128$ lattice, $\Delta x=1$ & RK4; fourth-order periodic finite difference Laplacian & $2\times10^{-3}$ &$\Delta t_{\mathrm{save}}=1$ \\
Brusselator & $[-1,1)^2$, $128\times128$ & Explicit Euler; second-order periodic central differences & $10^{-3}$ & $t\in[2,4]$ \\
Complex Ginzburg--Landau & $[-1,1)^2$, $128\times128$ & Explicit Euler; second-order periodic central differences & $10^{-3}$ & $t\in[2,10]$ \\
Schnakenberg & $[-1,1)^2$, $128\times128$ & Explicit Euler; second-order periodic central differences & $10^{-3}$ & $t\in[2,4]$ \\
Lambda--Omega & $[-1,1)^2$, $128\times128$ & Explicit Euler; second-order periodic central differences & $2\times10^{-3}$ & $t\in[3,12]$ \\
Barkley & Periodic square of side length $150$; $256\times 256\rightarrow128\times128$ & Strang splitting; explicit-midpoint reaction and exact spectral diffusion & $10^{-2}$ & $t\in[5,15]$ \\
Oregonator & $[-1,1)^2$, $128\times128$ & Explicit Euler; second-order periodic central differences & $2\times10^{-4}$ & $t\in[0.2,0.8]$ \\
\bottomrule
\end{tabularx}
\end{table}

After numerical integration, system-specific quality screening was applied before a trajectory was included in the benchmark. Simulations were excluded if they contained non-finite values, exhibited amplitude divergence, showed negligible spatial or temporal variation, or contained spatial structures that were not adequately resolved on the simulation grid. Gray--Scott, Brusselator, and Schnakenberg were screened using temporal-change criteria; complex Ginzburg--Landau was screened using spatial-variance and total-variation criteria; Oregonator was screened using temporal-change, concentration, and resolved-wavelength criteria; and Lambda--Omega was screened using a multi-frame change criterion. Barkley trajectories were screened for numerical validity.

\suppsubsection{S1.5 {Barkley benchmark with non-periodic boundary conditions}}
\label{supp:boundary_dataset}

{To assess adaptation beyond the periodic domains used for predictive pretraining, we generated three additional Barkley datasets on the same square domain of side length 150 as in the primary Barkley benchmark. Let $\Omega$ denote this domain and let $\partial_n u=\nabla u\cdot\mathbf{n}$ denote the outward normal derivative on $\partial\Omega$. For each dataset, the activator field $u$ satisfied one
of the following homogeneous boundary conditions: }
\begin{equation}
{
\left.
\begin{aligned}
\partial_n u &= 0 && \text{(Neumann)},\\
\partial_n u+\kappa u &= 0,\quad \kappa=0.15 && \text{(Robin)},\\
u &= 0 && \text{(Dirichlet)},
\end{aligned}
\right\}
\qquad \text{on }\partial\Omega .
}
\label{eq:supp_barkley_boundary_conditions}
\end{equation}
Because $D_v=0$, the inhibitor equation contains no spatial-diffusion term, and no spatial boundary operator is required for $v$. 

Each boundary-specific dataset comprised 20 adaptation trajectories, five validation trajectories, and 300 test trajectories. Every trajectory contained 40 stored two-channel fields. Simulations were performed on a cell-centered $256\times256$ grid, and the stored fields were subsequently average-pooled to $128\times128$. We used $\Delta t=0.01$ and retained 40 fields over the interval $t\in [5,15]$. The coefficient ranges were identical to those of the primary Barkley benchmark. 

For all three boundary conditions, the diffusion operator was approximated using a second-order five-point finite-difference Laplacian. Time integration used Strang splitting, with the reaction and diffusion subproblems advanced by the explicit midpoint method. The boundary conditions were imposed through second-order ghost-cell relations. Initial conditions were compact broken waves with a quiescent collar adjacent to the boundary and were therefore compatible with all three homogeneous boundary conditions. 

Before boundary-specific quality screening, common index-level randomization was used to align the coefficient draws and initial-condition parameters across the three boundary conditions. Because screening was performed separately for each condition, the retained datasets represent matched generating distributions but are not paired trajectory-by-trajectory. None of these trajectories was used during predictive pretraining. Within each boundary condition and support-set selection, all models used the same support trajectories and were evaluated on the same fixed test trajectories.

\suppnote{Supplementary Note 2: RD-JEPA architecture and predictive latent learning}
\label{supp:note_model}

\suppsubsection{S2.1 Overall two-stage RD-JEPA formulation}
\label{supp:jepa_formulation}
A joint-embedding predictive architecture (JEPA) learns to predict the latent representation of a target observation from an observed context~\cite{assran2023ijepa,bardes2024vjepa}. An encoder maps observations to learned feature vectors, referred to here as latent tokens. RD-JEPA applies this principle to reaction-diffusion trajectories in two stages. During predictive latent pretraining, an online encoder and a predictor are trained to predict target tokens at selected spatial patches and time offsets. During downstream adaptation, the online encoder is frozen, while the predictor is fine-tuned jointly with a decoder that produces future-field predictions from the predicted tokens and the latest observed field. 

\paragraph{Predictive latent pretraining.} 
Let \(x_t\in\mathbb{R}^{H\times W\times C}\) denote the discretized reaction-diffusion field at saved time index \(t\), expressed in the standardized coordinates. In all reported experiments, \(H=W=128\) and \(C=2\). The model receives a context window of four consecutive saved fields,
\begin{equation}
    X_t^c = \bigl(x_{t-c+1},\ldots,x_t\bigr) = \bigl(x_{t-3},x_{t-2},x_{t-1},x_t\bigr)
    \in \mathbb{R}^{c\times H\times W\times C}, \qquad c=4.
    \label{eq:supp_context}
\end{equation}
The online encoder \(E_\theta\) maps the context window to a spatial grid of latent tokens,
\begin{equation}
    Z_t^c = E_\theta(X_t^c) =
    \bigl(z_{t,1}^c,\ldots,z_{t,S}^c\bigr) \in \mathbb{R}^{S\times d_{\mathrm{lat}}}, \qquad\mbox{where } S=16\times16=256, \quad d_{\mathrm{lat}}=512.
    \label{eq:supp_context_latent}
\end{equation}
Each token \(z_{t,i}^c\in\mathbb{R}^{d_{\mathrm{lat}}}\) is associated with spatial patch \(i\) and aggregates spatial and temporal information from the context window. The encoder architecture is described in Section~\ref{supp:encoder}.

The target encoder \(E_\xi\) provides reference representations by processing each complete target field separately:
\begin{equation}
    Z_{t+h}^{\mathrm{tar}}
    = E_\xi(x_{t+h})
    \in \mathbb{R}^{S\times d_{\mathrm{lat}}}.
    \label{eq:supp-target-encoding}
\end{equation}
Its parameters \(\xi\) are updated via an exponential moving average (EMA) of the corresponding online encoder parameters. The target-encoder construction and EMA update are detailed in Section~\ref{supp:targets}.

For each pretraining sample, let
\(\mathcal I=((i_m,h_m))_{m=1}^{M}\)
be the ordered query list, where \(i_m\in\{1,\ldots,S\}\) identifies a spatial patch and \(h_m\) specifies a time offset. The predictor \(P_\phi\) processes the context tokens and all queries jointly, returning one predicted token per query:
\begin{equation}
    \widehat z_m
    = \bigl[P_\phi(Z_t^c,\mathcal I)\bigr]_m,
    \qquad
    z_m^{\mathrm{tar}}
    = \bigl[Z_{t+h_m}^{\mathrm{tar}}\bigr]_{i_m},
    \qquad m=1,\ldots,M.
    \label{eq:supp_jepa_prediction}
\end{equation}
Query selection and the predictor architecture are described in Sections~\ref{supp:targets} and~\ref{supp:predictor}, respectively. The online encoder and predictor are trained using the loss in Section~\ref{supp:pretraining}, which compares the normalized predicted and target tokens. A single RD-JEPA model is pretrained jointly on all five source systems.

\paragraph{Dense downstream forecasting.} 
During downstream adaptation, the online encoder parameters \(\theta\) are held fixed, while the predictor \(P_\phi\) and decoder \(D_\psi\) are trained jointly by updating \(\phi\) and \(\psi\). For a requested forecast offset \(h\), the complete query list
\(\mathcal I_h^{\mathrm{full}}=\bigl((i,h)\bigr)_{i=1}^{S}\)
requests a latent token at every spatial patch. The predictor, therefore, produces a complete latent grid,
\begin{equation}
    \widehat Z_{t+h}
    = P_\phi(Z_t^c,\mathcal I_h^{\mathrm{full}})
    \in \mathbb{R}^{S\times d_{\mathrm{lat}}}.
    \label{eq:supp-dense-latent-prediction}
\end{equation}
The decoder combines this predicted latent grid with the latest context field to produce
\begin{equation}
    \widehat x_{t+h}
    = D_\psi\left(\widehat Z_{t+h},x_t\right)
    \in \mathbb{R}^{H\times W\times C}.
    \label{eq:supp-overall-downstream-forecast}
\end{equation}
The field \(x_t\) is processed through a convolutional pathway at the original \(128\times128\) spatial resolution. We evaluate this construction for \(h\in\{1,2,3,4,5\}\). All five future fields are predicted directly from the same observed context, rather than by recursively feeding predicted fields back into the model.

\suppsubsection{S2.2 Online encoder \(E_\theta\): spatiotemporal
tokenization and multiscale spatial encoding}
\label{supp:encoder}
The online encoder \(E_\theta\) maps the four-frame context \(X_t^c\) to the spatial latent grid \(Z_t^c\) in three stages. It first tokenizes each frame independently, then aggregates the four tokens at each spatial position, and finally exchanges information across spatial positions through a multiscale encoder.

\paragraph{Per-frame tokenization.}
For each context frame \(r\in\{t-3,t-2,t-1,t\}\), let \(P_{r,i}=\operatorname{patch}_i(x_r) \in \mathbb R^{8\times8\times2}\) denote the \(i\)-th non-overlapping spatial patch, containing both scalar components of the reaction-diffusion field. Partitioning each \(128\times128\) field into these patches gives a \(16\times16\) token grid.

The two scalar components of each patch are projected separately, and the resulting feature vectors are averaged. Since both components are present in every patch, the combined operation can be represented as \(\Pi(P_{r,i})+e\), where \(\Pi:\mathbb R^{8\times8\times2}\to\mathbb R^{d_{\mathrm{lat}}}\) is a structured linear map and \(e\in\mathbb R^{d_{\mathrm{lat}}}\) collects the constant terms, including the averaged channel-identity embeddings. A residual two-layer mixing map then produces the patch token:
\begin{align}    \label{eq:supp_patch_token}
    y_{r,i} &= \Pi(P_{r,i})+e + W_2\sigma_{\mathrm{GELU}}\!\left(W_1\operatorname{LN}(\Pi(P_{r,i})+e)+b_1\right)+b_2.
\end{align}
Here, \(d_{\mathrm{lat}}=512\), \(W_1,W_2\in\mathbb R^{d_{\mathrm{lat}}\times d_{\mathrm{lat}}}\), and \(b_1,b_2\in\mathbb R^{d_{\mathrm{lat}}}\). The projection, offset, and mixing parameters are learned during pretraining and shared across spatial positions, context frames, samples, and source systems. The activation is \(\sigma_{\mathrm{GELU}}(s)=s\Phi(s)\), where \(\Phi\) is the standard normal cumulative distribution function. Layer normalization, denoted by \(\operatorname{LN}\), acts over the latent feature dimension, with \(10^{-5}\) added to the variance before taking its square root.

Spatial positions and relative frame indices are encoded by fixed sinusoidal embeddings. For an even integer \(D\), define
\[
    f_D(a)=
    \left[
        \bigl(\sin(a\,10000^{-2j/D})\bigr)_{j=0}^{D/2-1},
        \bigl(\cos(a\,10000^{-2j/D})\bigr)_{j=0}^{D/2-1}
    \right]^{\mathsf T} \in\mathbb R^D,
\]
with all sine components followed by all cosine components. Let \((a_i,b_i)\in\{0,\ldots,15\}^2\) be the row and column indices of patch \(i\), and let \(\ell(r)=r-(t-3)\in\{0,1,2,3\}\) be the position of frame \(r\) within the context window. The positional embeddings are
\begin{equation}
    e_i^{\mathrm{space}}=
    \begin{bmatrix}
        f_{d_{\mathrm{lat}}/2}(a_i)\\
        f_{d_{\mathrm{lat}}/2}(b_i)
    \end{bmatrix},
    \qquad e_{\ell(r)}^{\mathrm{time}} = f_{d_{\mathrm{lat}}}(\ell(r)),
    \label{eq:supp-positional-embeddings}
\end{equation}
both in \(\mathbb R^{d_{\mathrm{lat}}}\). The spatial embedding concatenates row and column embeddings, whereas the temporal embedding encodes the frame's relative position within the four-frame context.

During predictive pretraining, patch tokens may be masked after tokenization and before temporal aggregation. For the same-time target task (\(h=0\)), the tokens \(y_{t,i}\) at the requested target patches in the latest context field \(x_t\) are masked. The target-selection procedure is described in Section~\ref{supp:targets}.

Independent of the target type, each training sample receives additional random context masking with probability \(0.35\). For such a sample, a single masking probability \(p\) is drawn uniformly from \([0.05,0.25]\). Conditional on \(p\), each patch token in each of the four context frames is independently masked with probability \(p\). Thus, \(0.35\) controls whether a sample receives additional masking, whereas \(p\) controls the masking probability within that sample.

Let \(\mathcal M\) be the union of the same-time target mask and the additional random context mask, represented as pairs \((r,i)\) of frame and patch indices. Each selected token is replaced by a learned mask vector \(e_{\mathrm{enc}}^{\mathrm{mask}}\in \mathbb R^{d_{\mathrm{lat}}}\). The positional embeddings and a learned encoder embedding \(e_{\mathrm{enc}}\in\mathbb R^{d_{\mathrm{lat}}}\) are then added to obtain the input to temporal aggregation:
\begin{equation}
    \widetilde y_{r,i}=
    \begin{cases}
        e_{\mathrm{enc}}^{\mathrm{mask}},
        & (r,i)\in\mathcal M,\\[2mm]
        y_{r,i},
        & (r,i)\notin\mathcal M
    \end{cases}
    +e_i^{\mathrm{space}}
    +e_{\ell(r)}^{\mathrm{time}}
    +e_{\mathrm{enc}}.
    \label{eq:supp-context-token-embedding}
\end{equation}
Both \(e_{\mathrm{enc}}^{\mathrm{mask}}\) and \(e_{\mathrm{enc}}\) are shared across spatial positions, context frames, samples, and source systems. Masked tokens retain their spatial and temporal embeddings, so their locations and frame indices remain available to the encoder. The target encoder receives complete, unmasked target fields during pretraining. No context masking is applied during downstream forecasting.

\paragraph{Temporal aggregation.}
At each spatial position \(i\), the four embedded context tokens \(\widetilde y_{t-3,i},\ldots,\widetilde y_{t,i}\) are aggregated into a single vector \(\bar z_{t,i}\in\mathbb R^{d_{\mathrm{lat}}}\). A query is initialized as \(\widetilde y_{t,i}+b_{\mathrm{temp}}\), where \(b_{\mathrm{temp}}\in\mathbb R^{d_{\mathrm{lat}}}\) is learned, and is updated by three successive cross-attention blocks. In every block, the current query attends to the same four input tokens. These tokens supply the keys and values and are not themselves updated.

Each block applies an eight-head cross-attention update followed by a two-layer feedforward update, both with residual connections. The feedforward map uses GELU activation and intermediate dimension \(2d_{\mathrm{lat}}\). The three blocks have separate parameters, each shared across spatial positions.

The final queries form the temporally aggregated grid
\begin{equation}
    \overline Z_t^c
    =
    \bigl[\bar z_{t,1},\ldots,\bar z_{t,S}\bigr]^{\mathsf T}
    \in\mathbb R^{S\times d_{\mathrm{lat}}},
    \qquad S=256.
    \label{eq:supp-temporally-aggregated-grid}
\end{equation}
Temporal aggregation therefore reduces the four tokens at each patch position to one while preserving the \(16\times16\) spatial grid.

\paragraph{Multiscale spatial encoding.}
The temporally aggregated grid \(\overline Z_t^c\) is processed by a five-stage U-shaped spatial encoder with a resolution sequence
\begin{equation}
    16\times16
    \longrightarrow 8\times8
    \longrightarrow 4\times4
    \longrightarrow 8\times8
    \longrightarrow 16\times16.
    \label{eq:supp-encoder-scales}
\end{equation}
The stages contain \((2,3,6,3,2)\) operator blocks, respectively, and retain feature width \(d_{\mathrm{lat}}=512\) throughout. On the descending path, a \(3\times3\) convolution with stride \(2\) reduces the spatial resolution between successive stages. On the ascending path, bilinear interpolation followed by a \(3\times3\) convolution restores the resolution. At each ascending stage, the upsampled features are added to a learned projection of the descending-stage output at the same resolution, forming a U-shaped skip connection.

Each operator block applies three successive sub-operations to its input, each with a residual connection.

\textit{(i) Spatial self-attention.} An eight-head self-attention layer exchanges information across all tokens at the current resolution.

\textit{(ii) Spectral--local mixing.} Two branches process the same input in parallel. The \emph{spectral branch} applies a two-dimensional Fourier transform to the spatial grid of each feature component independently, multiplies a selected subset of Fourier coefficients (at most eight per spatial direction, subject to the resolution of the current stage) by learned complex weights, transforms the result back to the spatial domain, mapped back to $d_{\mathrm{lat}}$ components by a learned affine map. The \emph{local branch} applies a depthwise \(3\times3\) convolution (independently for each feature component), followed by GELU and a \(1\times1\) convolution. The outputs of the two branches are concatenated along the feature dimension and mapped back to $d_{\mathrm{lat}}$ components by a learned affine map.

\textit{(iii) Position-wise MLP.} The same two learned affine maps, separated by GELU and with intermediate width $4d_{\mathrm{lat}}$, are applied at every spatial position.

Before each of the three operations, layer normalization is applied separately to each token. The normalized components are then scaled and shifted using values computed from the shared encoder embedding $e_{\mathrm{enc}}$ by a learned nonlinear map. Each normalization layer has its own map; within that layer, the same scaling and shifting are applied at all spatial positions.

After the final stage, layer normalization is applied separately to the $512$ feature components of each token, giving
\begin{equation}
    Z_t^c
    =
    \bigl[z_{t,1}^c,\ldots,z_{t,S}^c\bigr]^{\mathsf T}
    \in\mathbb R^{S\times d_{\mathrm{lat}}},
    \qquad S=256.
    \label{eq:supp-context-encoder-output}
\end{equation}
Each output token remains associated with a spatial patch but incorporates information from the four context frames and other spatial positions. The resulting latent grid is supplied to the lead-time-conditioned predictor described in Section~\ref{supp:predictor}.

\suppsubsection{S2.3 Target sampling and EMA target encoding}
\label{supp:targets}

\paragraph{Spatial target region.}
For each pretraining sample, a source system is selected uniformly from the five source systems. Rectangular target regions are then sampled on the \(16\times16\) latent grid: the number of rectangles is drawn uniformly from \(\{1,\ldots,4\}\); each rectangle's height and width are drawn independently and uniformly from \(\{2,\ldots,6\}\) patches; and its upper-left corner is placed uniformly among all valid positions inside the grid. The union of all covered patch indices defines the spatial target region \(\mathcal{R}\subseteq\{1,\ldots,S\}\), with overlapping positions counted once.

\paragraph{Target type and query list.}
One target type is selected according to Table~\ref{tab:supp_target_sampling}, yielding a set of time offsets \(\mathcal{H}\). The candidate set \(\mathcal{C}=\mathcal{R}\times\mathcal{H}\) collects all patch--offset pairs at which the predictor must match the target encoder's output. Because \(|\mathcal{C}|\) varies across samples, a fixed number \(M=256\) of pairs is subsampled: if \(|\mathcal{C}|\ge M\), pairs are drawn without replacement; otherwise with replacement. This gives the ordered query list
\begin{equation}
    \mathcal I = \bigl((i_m,h_m)\bigr)_{m=1}^{M}, \qquad M=256.
    \label{eq:supp-target-index-set}
\end{equation}
\(M=256\) is the number of loss terms per training step; the number of masked patches is determined separately by \(|\mathcal{R}|\). Repeated pairs in \(\mathcal{I}\) occupy separate positions and contribute independently to the loss.

For the same-time target type (\(\mathcal{H}=\{0\}\)), the patches in \(\mathcal{R}\) are masked in the latest context frame \(x_t\) of the online encoder, as described in Section~\ref{supp:encoder}. For
future target types (\(h>0\)), no such masking is needed: future frames are not part of the online encoder input, so no information needs to be withheld.
\begin{table}[!ht]
\centering
\small
\caption{\textbf{Target selection during predictive pretraining.} All three target types use the same spatial-region sampling procedure to obtain \(\mathcal{R}\). The selected offset set \(\mathcal{H}\) defines the candidate set \(\mathcal{C}=\mathcal{R}\times\mathcal{H}\), from which \(M=256\) query pairs are subsampled for loss computation.}
\label{tab:supp_target_sampling}
\begin{tabularx}{\textwidth}
{@{}p{0.20\textwidth}p{0.14\textwidth}Y@{}}
\toprule
Target type & Probability & Selection of time offsets \\
\midrule
Future block
& 0.50
& \(\mathcal H=\{h\}\), where \(h\) is sampled uniformly from
  \(\{1,\ldots,8\}\). \\
Future tube
& 0.40
& \(\mathcal H\) contains four distinct offsets sampled uniformly
  without replacement from \(\{1,\ldots,8\}\). The same spatial
  region \(\mathcal R\) is used at each selected offset. \\
Same time
& 0.10
& \(\mathcal H=\{0\}\), targeting the latest context frame \(x_t\).
  The patches in \(\mathcal{R}\) are masked in the online encoder
  input. \\
\bottomrule
\end{tabularx}
\end{table}

\paragraph{Target encoding.}
For each distinct offset \(h\) appearing in \(\mathcal I\), the target encoder \(E_\xi\) processes the complete, unmasked field \(x_{t+h}\) as a separate single-frame input. The target encoder is a separate copy of the online encoder, with its own parameters \(\xi\) and the same parameterized modules for tokenization, temporal aggregation, and multiscale spatial encoding. The two encoders differ in temporal input length: the online encoder processes four frames, whereas the target encoder processes one. At each spatial position, the three temporal cross-attention blocks therefore attend to a single target-frame token. This change in sequence length does not alter the parameter dimensions, so the two encoders have matching parameter tensors.

The target frame is assigned relative temporal index \(0\) and uses \(e_0^{\mathrm{time}}\) for every prediction offset \(h\). This temporal index identifies the frame's position within the single-frame encoder input; it does not encode the prediction offset. The offset \(h\) is supplied to the latent predictor, not to the target encoder.

All patches remain visible throughout the target-encoder forward pass. The requested target vectors are selected from the resulting latent grids according to Equation~\eqref{eq:supp_jepa_prediction}. Thus, target sampling determines which output tokens enter the predictive latent loss, rather than which input patches are visible to the target encoder. The selected target vectors are treated as constants when differentiating this loss; no gradients are propagated through \(E_\xi\).

\paragraph{EMA update.}
The target encoder is initialized from the online encoder. Let \(s\) index pretraining steps. At step \(s\), the target encoder with parameters \(\xi_s\) provides the reference representations. After the optimizer updates the online-encoder parameters from \(\theta_s\) to \(\theta_{s+1}\), the target-encoder parameters are updated by
\begin{equation}
    \xi_0=\theta_0,
    \qquad
    \xi_{s+1}
    =m_s\xi_s+(1-m_s)\theta_{s+1}.
    \label{eq:supp-ema-update}
\end{equation}
The averaging coefficient \(m_s\) follows the cosine schedule. The target encoder is updated only through this EMA rule, not by the optimizer.

\suppsubsection{S2.4 Lead-time-conditioned predictor with diffusion-inspired and reaction-inspired branches}
\label{supp:predictor}
The latent predictor \(P_\phi\) receives the context representation \(Z_t^c\in\mathbb R^{S\times d_{\mathrm{lat}}}\) and the ordered query list \(\mathcal I=\bigl((i_m,h_m)\bigr)_{m=1}^{M}\). For each requested pair \((i_m,h_m)\), it predicts a latent vector \(\widehat z_m\in\mathbb R^{d_{\mathrm{lat}}}\) associated with spatial patch \(i_m\) at offset \(h_m\). The predictor initializes a query vector for each requested pair, using spatial and time-offset embeddings. Eight successive blocks then exchange information among the queries, incorporate the encoded context, and apply diffusion-inspired and reaction-inspired updates.

\paragraph{Encoding the requested time offset.}
For a requested offset \(h\in\{0,\ldots,8\}\), define \(\tau=h/8\) and construct
\begin{equation}
    \gamma(h)=
    \left[
        \bigl(\sin(2\pi 2^j\tau)\bigr)_{j=0}^{15},
        \bigl(\cos(2\pi 2^j\tau)\bigr)_{j=0}^{15},
        \tau,\tau^2
    \right]^{\mathsf T}
    \in\mathbb R^{34}.
    \label{eq:supp_time_features}
\end{equation}
A shared two-layer map produces the time-offset embedding
\begin{equation}
    e_h=
    \widetilde W_2
    \sigma_{\mathrm{GELU}}\!\left(
        \widetilde W_1\operatorname{LN}(\gamma(h))
        +\widetilde b_1
    \right)
    +\widetilde b_2
    \in\mathbb R^{d_{\mathrm{lat}}},
    \label{eq:supp_time_embedding}
\end{equation}
where \(\widetilde W_1\in\mathbb R^{2d_{\mathrm{lat}}\times34}\), \(\widetilde W_2\in \mathbb R^{d_{\mathrm{lat}}\times2d_{\mathrm{lat}}}\), \(\widetilde b_1\in\mathbb R^{2d_{\mathrm{lat}}}\), and \(\widetilde b_2\in\mathbb R^{d_{\mathrm{lat}}}\). Here, layer normalization is applied to the \(34\)-component input feature vector. All parameters of this map are shared across requested offsets.

\paragraph{Query initialization.}
For each requested pair \((i_m,h_m)\), the initial query is
\begin{equation}
    q_m^{(0)}
    =
    \operatorname{LN}\!\left(
        e_{i_m}^{\mathrm{grid}}
        +e_{i_m}^{\mathrm{space}}
        +e_{h_m}
        +e_{\mathrm{pred}}
    \right).
    \label{eq:supp_target_query}
\end{equation}
The vector \(e_i^{\mathrm{grid}}\in \mathbb R^{d_{\mathrm{lat}}}\) is a learned embedding specific to position \(i\) on the \(16\times16\) latent grid, whereas \(e_i^{\mathrm{space}}\) is the fixed sinusoidal spatial embedding defined in Section~\ref{supp:encoder}. The learned vector \(e_{\mathrm{pred}}\in\mathbb R^{d_{\mathrm{lat}}}\) is shared across samples, source systems, spatial positions, and requested offsets. These initial queries encode the requested locations and offsets; context information enters through the attention and diffusion-inspired operations described below.

The initial queries $q_m^{(0)}$ pass through eight successive blocks, each performing the complete sequence of attention, diffusion, reaction, and feedforward updates. Block $\ell\in\{0,\ldots,7\}$ takes $q_m^{(\ell)}$ as input and produces $q_m^{(\ell+1)}$.

\paragraph{Query and context attention.}

Each block first applies self-attention among the \(M\) queries, allowing requests at different spatial positions and time offsets to exchange information. It then applies context cross-attention: the current query vectors provide the queries, and the \(S=256\) context tokens in \(Z_t^c\) provide the keys and values. Thus, each requested vector can incorporate information from all spatial positions in the encoded context. Both attention modules use eight heads of dimension \(64\). Each head uses its own learned query, key, and value projections. The eight head outputs are concatenated and mapped back to \(\mathbb R^{d_{\mathrm{lat}}}\) by a learned output projection. Both attention modules use pre-normalization and residual connections.

\paragraph{Diffusion-inspired and reaction-inspired updates.}
For spatial patch \(i\), define the nearest-neighbor difference of the encoded context tokens by
\begin{equation}
    \Delta_{\mathrm{lat}}z_{t,i}^c
    =
    \frac{1}{4}
    \sum_{j\in N(i)}
    \left(z_{t,j}^c-z_{t,i}^c\right),
    \label{eq:supp_latent_laplacian}
\end{equation}
where \(N(i)\) contains the four nearest neighbors of patch \(i\) on the periodic \(16\times16\) latent grid.

For the \(m\)-th requested pair, let \(\tau_m=h_m/8\). The diffusion-inspired branch in block \(\ell\) produces
\begin{equation}
    d_m^{(\ell)}
    =
    F_{\mathrm{diff}}^{(\ell)}\!\left(
        \left[
            z_{t,i_m}^c,\,
            \Delta_{\mathrm{lat}}z_{t,i_m}^c,\,
            e_{h_m},\,
            \tau_m
        \right]
    \right),
    \label{eq:supp_diffusion_update}
\end{equation}
Let $\bar q_m^{(\ell)}$ denote the query after the self-attention and context cross-attention updates in block $\ell$. The diffusion correction $g_{\mathrm d}^{(\ell)}d_m^{(\ell)}$ is first added to this query. The reaction-inspired branch then produces
\begin{equation}
    r_m^{(\ell)}
    =
    F_{\mathrm{react}}^{(\ell)}\!\left(
        \left[
            \operatorname{LN}\!\left(
                \bar q_m^{(\ell)}
                +g_{\mathrm d}^{(\ell)}d_m^{(\ell)}
            \right),\,
            e_{h_m}
        \right]
    \right).
    \label{eq:supp_reaction_update}
\end{equation}
Here, brackets denote concatenation into a single input vector. After both corrections, the query is
\begin{equation}
    u_m^{(\ell)}
    =
    \bar q_m^{(\ell)}
    +g_{\mathrm d}^{(\ell)}d_m^{(\ell)}
    +g_{\mathrm r}^{(\ell)}r_m^{(\ell)}.
    \label{eq:supp_structured_fusion}
\end{equation}
A residual feedforward update then completes the block:
\begin{equation}
    q_m^{(\ell+1)}
    =
    u_m^{(\ell)}
    +
    F_{\mathrm{out}}^{(\ell)}\!\left(
        \operatorname{LN}\!\left(u_m^{(\ell)}\right)
    \right),
    \qquad m=1,\ldots,M.
    \label{eq:supp_predictor_update}
\end{equation}

Each of \(F_{\mathrm{diff}}^{(\ell)}\), \(F_{\mathrm{react}}^{(\ell)}\), and \(F_{\mathrm{out}}^{(\ell)}\) uses two affine maps separated by GELU, with output dimension \(d_{\mathrm{lat}}\). Their input dimensions are \(3d_{\mathrm{lat}}+1\), \(2d_{\mathrm{lat}}\), and \(d_{\mathrm{lat}}\), respectively, and their intermediate dimensions are \(2d_{\mathrm{lat}}\), \(4d_{\mathrm{lat}}\), and \(4d_{\mathrm{lat}}\).

Both $F_{\mathrm{diff}}^{(\ell)}$ and $F_{\mathrm{react}}^{(\ell)}$ begin with layer normalization of their complete concatenated inputs. The reaction branch also normalizes the query before concatenating it with $e_{h_m}$, as shown explicitly in Equation~\eqref{eq:supp_reaction_update}. Its two normalization layers therefore act on vectors of dimensions $d_{\mathrm{lat}}$ and $2d_{\mathrm{lat}}$, respectively, and have separate parameters.

The scalar weights \(g_{\mathrm d}^{(\ell)}\) and \(g_{\mathrm r}^{(\ell)}\) are shared across queries within each block, are unconstrained, and are initialized to \(0.25\). All eight blocks have separate attention, normalization, MLP, and scalar-weight parameters.

After the eighth block, the predictor returns
\begin{equation}
    P_\phi(Z_t^c,\mathcal I)
    =
    \bigl[\widehat z_1,\ldots,\widehat z_M\bigr]^{\mathsf T}
    \in\mathbb R^{M\times d_{\mathrm{lat}}},
    \qquad
    \widehat z_m=\operatorname{LN}\!\left(q_m^{(8)}\right).
    \label{eq:supp_predictor_output}
\end{equation}
The predicted and target vectors are normalized when computing the predictive latent objective described in Section~\ref{supp:pretraining}.

The two structured branches operate on learned latent representations rather than physical field values. They introduce neighborhood-based spatial differences and query-wise nonlinear updates, but are not numerical discretizations of the physical diffusion and reaction operators. In particular, the reaction-inspired map acts separately on each query, although that query already contains information from other positions and offsets through attention. The same periodic neighbor sets \(N(i)\) are used in all experiments, including the Barkley experiments with non-periodic physical boundary conditions.

\suppsubsection{S2.5 Predictive latent objective and pretraining}
\label{supp:pretraining}
\paragraph{Predictive latent objective.}
Before computing the loss, each predicted and target vector is normalized independently across its latent feature components by subtracting its mean and dividing by the square root of its variance plus $10^{-5}$. 
For a batch of \(B\) samples, each containing \(M=256\) requested patch-offset pairs, the predictive latent objective is
\begin{equation}
    \mathcal L_{\mathrm{JEPA}}
    =
    \frac{1}{BMd_{\mathrm{lat}}}
    \sum_{b=1}^{B}\sum_{m=1}^{M}
    \left\|
        LN(\widehat z_{b,m})
        -
        LN(z_{b,m}^{\mathrm{tar}})
    \right\|_2^2.
    \label{eq:supp-jepa-loss}
\end{equation}
Here, \(\widehat z_{b,m}\) and \(z_{b,m}^{\mathrm{tar}}\) are the predicted and target vectors for the \(m\)-th query in sample \(b\). The loss averages the squared differences over samples, query positions, and latent feature components. The index \(m\) runs over positions in the ordered query list, so repeated patch--offset pairs contribute separately.

\paragraph{Optimization and EMA updates.}
Gradients of \(\mathcal L_{\mathrm{JEPA}}\) propagate through the normalization of the predicted vectors and update the online-encoder parameters \(\theta\) and predictor parameters \(\phi\). The target vectors are treated as constants when differentiating the loss; the target encoder receives neither gradient nor optimizer updates. After each optimizer step, the target-encoder parameters \(\xi\) are updated by the EMA rule in Equation~\eqref{eq:supp-ema-update}, using the newly updated online-encoder parameters \(\theta\).

Predictive pretraining runs for \(200{,}000\) optimization iterations. The EMA coefficient \(m_s\) increases from \(0.996\) to \(0.99995\) according to a cosine schedule over these iterations. The complete architecture and optimization settings are summarized in Tables~\ref{tab:supp_jepa_architecture} and~\ref{tab:supp_optimization}.

\suppsubsection{S2.6 Dense forecasting and adaptation from few trajectories}
\label{supp:adaptation}

\paragraph{Prediction on the full grid.}
During downstream adaptation, the target encoder is discarded, and the pretrained online encoder parameters \(\theta\) are held fixed. The pretrained predictor parameters \(\phi\) are fine-tuned jointly with the decoder parameters \(\psi\), which are initialized randomly. Context masking is disabled.

For a forecast offset \(h\), the predictor receives the complete query list \(\mathcal I_h^{\mathrm{full}}\) defined in Section~\ref{supp:jepa_formulation} and produces
\begin{equation}
    \widehat Z_{t+h}
    =
    P_\phi\!\left(
        E_\theta(X_t^c),\mathcal I_h^{\mathrm{full}}
    \right)
    \in\mathbb R^{S\times d_{\mathrm{lat}}}.
    \label{eq:supp_downstream_latent_grid}
\end{equation}
The decoder combines this predicted latent grid with the latest context field to produce
\begin{equation}
    \widehat x_{t+h} =
    D_\psi\!\left(\widehat Z_{t+h},x_t\right)
    \in\mathbb R^{H\times W\times C},
    \qquad h\in\{1,2,3,4,5\}.
    \label{eq:supp_direct_forecast}
\end{equation}
For each offset, the predictor processes all \(S=256\) spatial queries jointly. Different offsets are evaluated using separate query lists, with the same encoder, predictor, and decoder parameters. All five future fields are predicted directly from the same four-frame context; predicted fields are not fed back into the model.

The decoder is a U-shaped convolutional network. It combines the predicted \(16\times16\) latent grid with a multiscale feature pyramid extracted from \(x_t\) and reconstructs a two-channel field on the original \(128\times128\) grid.

\paragraph{Features extracted from the latest context field.}
The decoder constructs four feature arrays from \(x_t\):
\[
    \begin{array}{c|cccc}
        \text{array}
        & F_0 & F_1 & F_2 & F_3 \\ \hline
        \text{spatial resolution}
        & 128\times128
        & 64\times64
        & 32\times32
        & 16\times16 \\
        \text{feature width}
        & 96 & 192 & 384 & 384
    \end{array}.
\]
A residual convolutional block maps the two-channel input field to \(F_0\). For \(k=1,2,3\), a \(4\times4\) convolution with stride \(2\) and zero padding of width \(1\), followed by a residual convolutional block, maps \(F_{k-1}\) to \(F_k\). This feature pyramid depends only on \(x_t\) and is shared across the five forecast offsets.

Each residual block contains a branch with two successive \(3\times3\) convolutions, each followed by group normalization and the SiLU activation. The output of this branch is added to the block input. When the input and output feature widths differ, a \(1\times1\) convolution is applied to the skip path before the addition.

Group normalization divides the feature channels into eight groups. For each sample, normalization statistics are computed over the channels and spatial positions within each group, followed by learned per-channel scales and shifts. The SiLU activation is \(\operatorname{SiLU}(s) = \frac{s}{1+\exp(-s)}.\) Unless otherwise stated, all \(3\times3\) decoder convolutions use stride \(1\) and zero padding of width \(1\), preserving the spatial resolution.

\paragraph{Recovering the field on the original grid.} 
Each predicted token in \(\widehat Z_{t+h}\) is layer-normalized across its feature components and mapped from \(d_{\mathrm{lat}}=512\) to \(384\) components by a learned affine map shared across spatial positions and forecast offsets. The resulting \(16\times16\) feature array is concatenated with \(F_3\) along the feature dimension. A residual block maps these \(768\)-component features to \(384\) components.

Three successive decoding stages then double the spatial resolution by bilinear interpolation. At resolutions \(32\times32\), \(64\times64\), and \(128\times128\), the interpolated decoder features are concatenated with \(F_2\), \(F_1\), and \(F_0\), respectively. The concatenated feature widths are \(768\), \(576\), and \(288\); residual blocks reduce them to \(384\), \(192\), and \(96\), respectively.

A final \(3\times3\) convolution preserves the width \(96\) and is followed by group normalization and SiLU. A \(1\times1\) convolution then produces the two components of the predicted future field in standardized coordinates.

\paragraph{Standardized and original field coordinates.}
For each system, the mean $\mu_j$ and standard deviation $\sigma_j$ of physical component $j\in\{1,2\}$ are computed over all trajectories, time frames, and spatial positions in its pretraining dataset. The same statistics are used to standardize context fields and training targets. The standardization and inverse transformation are
\begin{equation}
    \sigma_j^* = \max(\sigma_j,10^{-6}).
    \qquad x_{t,j} = \frac{x_{t,j}^{\mathrm{orig}}-\mu_j}{\sigma_j^*},
    \qquad \widehat x_{t+h,j}^{\mathrm{orig}} = \sigma_j^*\widehat x_{t+h,j}+\mu_j
    \label{eq:supp_field_coordinates}
\end{equation}
The superscript $\mathrm{orig}$ denotes the original field scale.
The same constants are used for context fields and training targets.
Training losses are computed on standardized fields, whereas
reported forecasting errors are computed on the original scale.

\paragraph{Adaptation protocol.}
A separate predictor-decoder pair is adapted for each system, trajectory budget \(K\), and support-set selection, where the support set consists of the \(K\) training trajectories available for adaptation. For source systems included in pretraining, \(K\in\{5,10,20\}\). For systems excluded from pretraining, \(K\in\{1,5,10\}\). Each RD-JEPA adaptation run in the primary benchmark uses \(5{,}000\) optimization steps.

\paragraph{Downstream objective.}
The downstream objective gives equal weight to the five forecast offsets:
\begin{equation}
    \mathcal L_{\mathrm{down}} = \frac{1}{5}\sum_{h=1}^{5}
    \left(
        \mathcal L_{\mathrm{MSE}}^{(h)}
        +0.50\mathcal L_{\mathrm{rel}}^{(h)}
        +0.10\mathcal L_{\nabla}^{(h)}
        +0.05\mathcal L_{\mathrm{FFT}}^{(h)}
    \right).
    \label{eq:supp_downstream_loss}
\end{equation}
For a fixed offset, let \(u=(u_b)_{b=1}^{B}\) and \(\widehat u=(\widehat u_b)_{b=1}^{B}\) denote a batch of standardized target and predicted fields, respectively, with \(u_b,\widehat u_b\in\mathbb R^{H\times W\times C}\). Suppressing the offset superscript, the four loss terms are
\begin{align}
    \mathcal L_{\mathrm{MSE}}
    &=
    \left\langle(\widehat u-u)^2\right\rangle,
    \label{eq:supp_training_mse}
    \\
    \mathcal L_{\mathrm{rel}}
    &=
    \frac{1}{B}\sum_{b=1}^{B}
    \frac{
        \|\widehat u_b-u_b\|_2
    }{
        \max(\|u_b\|_2,10^{-8})
    },
    \label{eq:supp_training_relative}
    \\
    \mathcal L_{\nabla}
    &=
    \left\langle
        \left|\delta_x\widehat u-\delta_xu\right|
    \right\rangle
    +
    \left\langle
        \left|\delta_y\widehat u-\delta_yu\right|
    \right\rangle,
    \label{eq:supp_training_first_difference}
    \\
    \mathcal L_{\mathrm{FFT}}
    &=
    \left\langle
        \left|
            \log\!\left(
                1+|\mathcal F(\widehat u)|
            \right)
            -
            \log\!\left(
                {\color{red}1+}|\mathcal F(u)|
            \right)
        \right|
    \right\rangle.
    \label{eq:supp_training_fft}
\end{align}
Here, \(\langle\cdot\rangle\) denotes the mean over all entries of its argument, including the batch and both field components. Squares and absolute values are applied entrywise. The norm \(\|\cdot\|_2\) is the Euclidean norm over all spatial values and both components of one sample.

The operators \(\delta_x\) and \(\delta_y\) take adjacent first differences along the two spatial directions. They do not divide by the grid spacing and do not include wrap-around differences at the boundaries. The directional averages in Equation~\eqref{eq:supp_training_first_difference} are computed separately and then added.

The transform \(\mathcal F\) is a two-dimensional real-input Fourier transform applied independently to each field component, with coefficient normalization \(1/\sqrt{HW}\). For \(H=W=128\), it returns \(128\times65\) complex coefficients, retaining the nonnegative frequencies in the second spatial direction. All returned Fourier magnitudes receive equal weight in Equation~\eqref{eq:supp_training_fft}; no additional multiplicity weights are applied to account for omitted conjugate frequencies.

The no-predictive-latent control uses the same downstream objective. Objectives and optimization settings for the external supervised baselines are specified separately in Supplementary Section~S3.3 and Table~\ref{tab:supp_optimization}.

\suppsubsection{S2.7 Architecture and optimization summary}
\label{supp:model_summary}
Table~\ref{tab:supp_jepa_architecture} summarizes the RD-JEPA architecture and parameter counts. Table~\ref{tab:supp_optimization} lists the optimization settings
for predictive pretraining and downstream adaptation.
\begin{table}[!ht]
\centering
\small
\caption{\textbf{RD-JEPA input and output dimensions and parameter counts.} Shapes omit the batch dimension. The decoder output corresponds to one forecast offset. Parameter counts are reported in millions (M). Totals are computed before rounding the displayed component counts. The pretraining total includes the EMA target encoder.}
\label{tab:supp_jepa_architecture}

\begin{tabularx}{\textwidth}
{@{}p{0.24\textwidth}YY@{}}
\toprule
\textbf{Component} & \textbf{Input} & \textbf{Output} \\
\midrule
Online encoder
& $4\times128\times128\times2$
& $256\times512$ \\

Target encoder
& $1\times128\times128\times2$
& $256\times512$ \\

Latent predictor
& $256\times512$ context tokens and $256$ query pairs $(i,h)$
& $256\times512$ predicted tokens \\

Forecast decoder
& $256\times512$ predicted tokens and the latest
  $128\times128\times2$ context field
& $128\times128\times2$ \\
\bottomrule
\end{tabularx}

\medskip

\begin{tabularx}{\textwidth}
{@{}p{0.40\textwidth}Y@{}}
\toprule
\textbf{Parameter count} & \textbf{Value} \\
\midrule
Online encoder parameters
& 110.48M \\

EMA target encoder parameters
& 110.48M \\

Latent predictor parameters
& 76.37M \\

Forecast decoder parameters
& 20.54M \\

Total pretraining parameters
& 297.33M: online encoder, predictor, and EMA target encoder \\

Trainable parameters during pretraining
& 186.85M: online encoder and predictor \\

Trainable parameters during adaptation
& 96.90M: predictor and decoder \\
\bottomrule
\end{tabularx}
\end{table}

\begin{table}[t]
\centering
\small
\caption{\textbf{Optimization settings for predictive pretraining and downstream adaptation.} Each training iteration uses one batch. The downstream settings apply to RD-JEPA adaptation runs in the primary benchmark, for both source systems and systems excluded from pretraining.}
\label{tab:supp_optimization}

\begin{tabularx}{\textwidth}{@{}p{0.32\textwidth}Y@{}}
\toprule
\textbf{Setting} & \textbf{Value} \\
\midrule

\multicolumn{2}{@{}l}{\textbf{Shared settings}} \\
\addlinespace[2pt]
Optimizer
& AdamW, $(\beta_1,\beta_2)=(0.9,0.95)$ \\
Batch size
& 4 \\
Gradient clipping
& Global norm 1.0 \\

\midrule
\multicolumn{2}{@{}l}{\textbf{Predictive pretraining}} \\
\addlinespace[2pt]
Trainable modules
& Online encoder and predictor \\
Encoder updates
& Online encoder: optimizer; target encoder: EMA \\
Training iterations
& $200{,}000$ \\
Learning rate
& $7\times10^{-5}$ peak; $10^{-6}$ minimum \\
Learning rate schedule
& Linear warmup over the first $5{,}000$ iterations,
  followed by cosine decay \\
Weight decay
& 0.05 \\

\midrule
\multicolumn{2}{@{}l}{\textbf{Downstream adaptation}} \\
\addlinespace[2pt]
Trainable modules
& Predictor and decoder \\
Encoder updates
& Online encoder: frozen; target encoder: not used \\
Training iterations
& $5{,}000$ \\
Learning rate
& $10^{-5}$ predictor; $2\times10^{-4}$ decoder \\
Learning rate schedule
& Constant learning rates \\
Weight decay
& $10^{-4}$ \\

\bottomrule
\end{tabularx}
\end{table}

\suppnote{Supplementary Note 3: Experimental protocols, baselines and statistics}
\label{supp:note_experiments}

\suppsubsection{S3.1 Forecasting tasks, support selection and evaluation}
\label{supp:forecasting_protocol}

Each forecasting task is defined by four consecutive observed two-channel fields and five target fields at lead times \(h\in\{1,\ldots,5\}\), measured in stored-frame intervals. Predictions are obtained directly, without using earlier predicted fields as inputs. The no-predictive-latent control uses only the last of the four observed fields and the requested lead time, as described in Supplementary Section~\ref{supp:baseline_models}. For each system, the selected \(K\) training trajectories form the support set. We use \(K\in\{5,10,20\}\) for the five systems included in pretraining and \(K\in\{1,5,10\}\) for Lambda-Omega, Barkley, and Oregonator, which were excluded from pretraining. A separate model is trained for each system, value of \(K\), and support-set selection.

Three prespecified random seeds, 777, 778, and 779, are used to permute the trajectory indices in each primary-benchmark adaptation pool. For each seed and value of \(K\), the first \(K\) trajectories in the permutation form the support set; thus, each smaller support set is contained in the larger sets obtained with the same seed. All methods use the same selected trajectories for a given system, seed, and value of \(K\). For each method, system, and value of \(K\), the three runs vary only the support-set selection; the model-training seed remains fixed at 777. All RD-JEPA runs start from the same pretrained parameter values. The data partitions and test trajectories remain fixed across runs.

For each system and physical field component, all methods use the same fixed affine transformation, consisting of a scaling and a shift, as defined in Equation~\eqref{eq:supp_field_coordinates}. Its coefficients are estimated without using test trajectories. Errors are computed from predictions and reference fields on the original field scale.

For each source system, evaluation uses 300 test trajectories: 100 in-distribution, 100 coefficient-out-of-distribution, and 100 initial-condition-out-of-distribution trajectories. For each system excluded from pretraining, all runs use the same 300 trajectories from that system's in-distribution test set. Every valid temporal window is evaluated. With 40 stored frames, four context frames, and a maximum lead time of five, each trajectory contributes \(40-4-5+1=32\) windows. All methods, including the no-predictive-latent control, are evaluated on these same windows at the five prescribed lead times. In both primary benchmarks, each model is trained for \(5{,}000\) parameter updates.

In the boundary-condition experiment, support sets containing \(K\in\{1,5,10\}\) trajectories are selected separately for each boundary condition from a pool of 20 adaptation trajectories. Three distinct selections are obtained using random seeds 777, 778, and 780. Seed 779 is excluded because, under the fixed selection procedure, it selects the same \(K=1\) support trajectory as seed 778. The model-training seed remains fixed at 777. Within each boundary condition and support-set selection, all models use the same selected training trajectories and are evaluated on the same fixed set of 300 test trajectories. The number of training updates used by each model in this boundary-condition experiment is reported in Supplementary Note~\ref{supp:note_robustness}.

\suppsubsection{S3.2 {Comparison models and controls}}
\label{supp:baseline_models}

The comparison includes spectral neural operators, a rotationally equivariant transformer, a convolutional encoder--decoder, and two controls. Each supervised baseline maps four two-channel context fields directly to five future fields and is trained separately for every equation, value of $K$, and support-set selection.

\paragraph{Fourier Neural Operator.}\cite{li2021fourier} FNO uses six Fourier layers of width 192 with $16\times16$ retained modes, spatial coordinates, and a joint five-horizon output head.

\paragraph{Laplace Neural Operator.}\cite{cao2024laplace} LNO uses width 48, one pole-residue layer, and four retained pole modes along each spatial direction.

\paragraph{Riesz Neural Operator.}\cite{liu2026rieszno} RieszNO uses width 36, one spectral layer, $8\times8$ retained modes, spatial coordinates, and a 128-dimensional readout.

\paragraph{ReViT.}\cite{wei2026revit} The ReViT configuration uses feature widths ranging from
192 to 768, stage depths $(2,4,8,4,2)$, 16 attention heads, window size 8, and a direct five-horizon decoder. Local spatial gradients define the reference vectors used to construct invariant tokens.

\paragraph{ConvNeXt U-Net.}\cite{ohana2024the} CNextU-Net uses four encoder--decoder scales, two ConvNeXt blocks per scale, one bottleneck block and widths $42\rightarrow84\rightarrow168\rightarrow336\rightarrow672$.

\paragraph{No-predictive-latent control.} This model is trained independently with the JEPA encoder--predictor pathway removed. It retains the dense decoder and full-resolution last-frame pathway used by \modelname{}, receives a continuous lead-time embedding, and does not construct a sample-specific predictive latent grid. The comparison evaluates the contribution of the complete sample-dependent predictive-latent pathway.

\paragraph{Architecture-matched model trained from scratch.}
To compare source-pretrained adaptation with end-to-end supervised training from scratch, this control uses the same encoder-predictor-decoder architecture as RD-JEPA but initializes all three modules randomly. It is trained on Lambda--Omega, Barkley, and Oregonator, the three systems excluded from pretraining. Unlike RD-JEPA adaptation, which freezes the online encoder, this control jointly updates the encoder, predictor, and decoder.

The control uses the same support trajectories, direct forecast horizons, downstream objective, update budget, and fixed model-training random-number-generator state as RD-JEPA. No pretrained tensor, EMA target branch, or JEPA objective is used during this training. The model is trained for \(5{,}000\) updates with AdamW, using learning rates of \(10^{-4}\) for the encoder and predictor and \(2\times10^{-4}\) for the decoder. The final update checkpoint is evaluated; neither validation nor test results are used to select the checkpoint.

\suppsubsection{S3.3 Training objectives and optimization settings}
\label{supp:baseline_optimization}

All models, including \modelname{}, the no-predictive-latent control, the architecture-matched model trained from random initialization, and the five comparison models, use the loss in \cref{eq:supp_downstream_loss}. The weights of $\mathcal{L}_{\mathrm{MSE}}$, $\mathcal{L}_{\mathrm{rel}}$, $\mathcal{L}_{\nabla}$ and $\mathcal{L}_{\mathrm{FFT}}$ are $1$, $0.50$, $0.10$ and $0.05$, respectively, for every model.

The losses at the five forecast lead times receive equal weight. In \modelname{}, the pretrained predictor and the newly initialized decoder use different learning rates. The five comparison models are trained from random initialization using the same optimizer settings. These settings are listed in \cref{tab:supp_baseline_optimization}; the settings for the architecture-matched model trained from random initialization are given in S3.2.

All \modelname{} runs start from the same pretrained parameter values, obtained with pretraining seed 1234. For each model, the random-number generator used for training is initialized with seed 777 at the start of every run. All models use the same selected training trajectories within each support-set selection. Across the three selections, the support trajectories change while the pretraining and model-training seeds remain fixed. The reported variation therefore measures sensitivity to support-set selection; it does not assess variability across independently chosen training or pretraining seeds.

For every model in the primary benchmark, evaluation uses the parameter values obtained after 5,000 training updates. Training does not use early stopping, and validation results are not used to select the parameter values evaluated.

\begin{table}[t]
\centering
\footnotesize
\caption{\textbf{Downstream optimization settings for the primary benchmark.} The external-baseline column applies to FNO, LNO, RieszNO, ReViT and CNextU-Net.}
\label{tab:supp_baseline_optimization}
\begin{tabularx}{\textwidth}{@{}p{0.22\textwidth}p{0.22\textwidth}p{0.22\textwidth}Y@{}}
\toprule
\textbf{Setting} & \textbf{RD-JEPA} & \textbf{No predictive latent} & \textbf{External baselines} \\
\midrule
Training updates & 5,000 & 5,000 & 5,000 \\
Batch size & 4 & 4 & 4 \\
Optimizer & AdamW & AdamW & AdamW \\
Learning rate & $10^{-5}$ predictor; $2\times10^{-4}$ decoder & $10^{-4}$ conditioner; $2\times10^{-4}$ decoder & $2\times10^{-4}$ \\
Weight decay & $10^{-4}$ & $10^{-4}$ & $10^{-4}$ \\
Adam betas & $(0.9,0.95)$ & $(0.9,0.95)$ & $(0.9,0.95)$ \\
Gradient clipping & 1.0 & 1.0 & 1.0 \\
Weight of $\mathcal{L}_{\mathrm{rel}}$ & 0.50 & 0.50 & 0.50 \\
Weight of $\mathcal{L}_{\nabla}$ & 0.10 & 0.10 & 0.10 \\
Weight of $\mathcal{L}_{\mathrm{FFT}}$ & 0.05 & 0.05 & 0.05 \\
{Support selection and fixed training seed} & \multicolumn{3}{>{\raggedright\arraybackslash}p{0.68\textwidth}}{Primary benchmark support-selection seeds: 777, 778 and 779; model-training seed: 777 for every support selection; one fixed RD-JEPA predictive-pretraining checkpoint (pretraining seed 1234)} \\
\bottomrule
\end{tabularx}
\end{table}

\suppsubsection{S3.4 Evaluation metrics and statistical aggregation}
\label{supp:evaluation_metrics}

{All errors are evaluated in the original physical field units.} For prediction $\widehat{\mathbf{x}}_i^{(h)}$ and reference field $\mathbf{x}_i^{(h)}$, the relative field error is
\begin{equation}
E_{\mathrm{rel}L^2,i}^{(h)}=\frac{\left\|\widehat{\mathbf{x}}_i^{(h)}-\mathbf{x}_i^{(h)}\right\|_2}{\left\|\mathbf{x}_i^{(h)}\right\|_2+\epsilon},
\qquad \epsilon=10^{-8},
\label{eq:supp_metric_rel_l2}
\end{equation}
and the spatial-gradient error is
\begin{equation}
E_{\nabla L^1,i}^{(h)}=\operatorname{MAE}\!\left(\delta_x\widehat{\mathbf{x}}_i^{(h)},\delta_x\mathbf{x}_i^{(h)}\right)+\operatorname{MAE}\!\left(\delta_y\widehat{\mathbf{x}}_i^{(h)},\delta_y\mathbf{x}_i^{(h)}\right).
\label{eq:supp_metric_gradient}
\end{equation}
{Here, $\delta_x$ and $\delta_y$ are adjacent first differences along the two grid axes, without a wrap-around term. Each mean absolute error is averaged over both field channels and all valid neighbouring grid pairs. The same definition is used for periodic and non-periodic datasets.}
{Relative $L^2$ and gradient $L^1$ are the principal reported metrics. MSE is retained as a secondary diagnostic in Supplementary Data~1--3.}

{For support-selection run $r$, errors are first averaged over all valid windows within each test trajectory. Let $e_{p,m,K,h,i,w}^{(r)}$ denote the error for window $w$ of trajectory $i$. The trajectory-level value is}
\begin{equation}
{
E_{p,m,K,h,i}^{(r)}
=
\frac{1}{W_i}\sum_{w=1}^{W_i}e_{p,m,K,h,i,w}^{(r)},
\qquad W_i=32 .
}
\label{eq:supp_trajectory_mean}
\end{equation}
{The run-level mean for equation $p$, model $m$, trajectory budget $K$ and horizon $h$ is}
\begin{equation}
{
\overline{E}_{p,m,K,h}^{(r)}
=
\frac{1}{N_p}\sum_{i=1}^{N_p}E_{p,m,K,h,i}^{(r)},
\qquad N_p=300 .
}
\label{eq:supp_per_horizon_mean}
\end{equation}
{The horizon-averaged value for each repeat is}
\begin{equation}
{
\overline{E}_{p,m,K}^{(r)}
=
\frac{1}{5}\sum_{h=1}^{5}\overline{E}_{p,m,K,h}^{(r)} .
}
\label{eq:supp_horizon_macro_average}
\end{equation}
{For every source- and held-out-equation analysis, $R=3$ denotes the three support-set selections. The final mean and sample standard deviation are computed across the $R$ run-level means,}
\begin{equation}
{
\mu_{p,m,K,h}=\frac{1}{R}\sum_{r=1}^{R}\overline{E}_{p,m,K,h}^{(r)},
\qquad
s_{p,m,K,h}=\sqrt{\frac{1}{R-1}\sum_{r=1}^{R}\left(\overline{E}_{p,m,K,h}^{(r)}-\mu_{p,m,K,h}\right)^2} .
}
\label{eq:supp_run_aggregation}
\end{equation}
{Thus, $n=3$ denotes the three support-set selections. Each run-level value summarizes the same fixed set of 300 test trajectories. The reported sample standard deviation is computed exclusively across the three run-level means and therefore measures support-set sensitivity; heterogeneity across test trajectories is not used as the uncertainty bar. For the split-resolved source analysis, $N_p=100$ within each test regime before averaging equations and horizons.}

{Percentage reductions are paired by support-set selection. For baseline $b$ and metric $E$, the reduction in run $r$ is}
\begin{equation}
{
Q_{p,b,K}^{(r)}
=100\left(1-\frac{\overline{E}_{p,\mathrm{RD\mbox{-}JEPA},K}^{(r)}}{\overline{E}_{p,b,K}^{(r)}}\right),
}
\label{eq:supp_paired_reduction}
\end{equation}
{and displayed centres and error bars are the mean and sample standard deviation of the three paired values $Q_{p,b,K}^{(r)}$. This is a mean of matched run-level reductions, not a ratio formed after pooling the runs. Supplementary Data~1 contains the source-equation run-level values, Supplementary Data~2 contains the held-out-equation horizon-resolved run-level values, and Supplementary Data~3 contains the broad-comparison horizon-averaged run-level values. {Supplementary Data~4 contains the architecture-matched scratch run-level values, and Supplementary Data~5 contains the boundary-condition stress-test run-level values.}}

\suppnote{Supplementary Note 4: Complete forecasting results on source systems}
\label{supp:note_source_results}

\suppsubsection{S4.1 Evaluation settings and error averaging}
\label{supp:source_result_aggregation}

We compare \modelname{}, the no-predictive-latent control and FNO on the five source systems, using $K\in\{5,10,20\}$ training trajectories and forecast lead times $h=1,\ldots,5$. For each support-set selection, errors are first averaged over valid windows within each trajectory and then over the same 300 test trajectories, comprising 100 trajectories from each of the three test regimes. The reported means and sample standard deviations are computed across the three support-set selections. Metric definitions and averaging procedures are given in Supplementary Note~\ref{supp:note_experiments}, and the values for individual selections are provided in Supplementary Data~1.

After combining the three test regimes, \modelname{} has the lowest mean error among the three models on both reported metrics for all $5\times3\times5=75$ combinations of source system, support-set size and forecast lead time.

\suppsubsection{S4.2 Complete relative-field-error results}
\label{supp:source_rel_results}

The relative $L^2$ errors defined in \cref{eq:supp_metric_rel_l2} are reported for $K=5$, $10$ and $20$ in \cref{tab:supp_source_rel_k5,tab:supp_source_rel_k10,tab:supp_source_rel_k20}, 
respectively. Each table gives the error at each forecast lead time and its average over $h=1,\ldots,5$.

\begin{table}[!htbp]
\centering
\scriptsize
\caption{{\textbf{Complete relative $L^2$ results on the five source systems at $K=5$.} Entries are $100\times$ the error and report the mean $\pm$ sample s.d. across three paired support-set selections. Within each run, errors are first averaged over all valid windows within each trajectory and then over the same 300 test trajectories (100 from each test regime). The Avg. column averages $h=1,\ldots,5$ within each run before across-run aggregation. Lower values are better. {Boldface marks the lowest numerical mean within each equation and horizon; no formal hypothesis tests were performed.}}}
\label{tab:supp_source_rel_k5}
\resizebox{\textwidth}{!}{%
\begin{tabular}{llcccccc}
\toprule
\textbf{PDE} & \textbf{Model} & \textbf{$h=1$} & \textbf{$h=2$} & \textbf{$h=3$} & \textbf{$h=4$} & \textbf{$h=5$} & \textbf{Avg.} \\
\midrule
\multirow{3}{*}{Gray--Scott} & \modelname{} & \sbestmeanstd{3.56}{1.21} & \sbestmeanstd{4.66}{0.97} & \sbestmeanstd{5.74}{0.84} & \sbestmeanstd{6.83}{0.75} & \sbestmeanstd{7.99}{0.66} & \sbestmeanstd{5.76}{0.87} \\
 & No predictive latent & \smeanstd{6.54}{1.68} & \smeanstd{10.13}{2.38} & \smeanstd{13.40}{2.97} & \smeanstd{16.35}{3.19} & \smeanstd{19.00}{2.85} & \smeanstd{13.08}{2.58} \\
 & FNO & \smeanstd{6.10}{0.49} & \smeanstd{7.30}{0.22} & \smeanstd{9.36}{0.56} & \smeanstd{11.55}{0.93} & \smeanstd{13.71}{1.29} & \smeanstd{9.60}{0.46} \\
\midrule
\multirow{3}{*}{FitzHugh--Nagumo} & \modelname{} & \sbestmeanstd{4.11}{1.38} & \sbestmeanstd{4.59}{1.34} & \sbestmeanstd{5.36}{1.20} & \sbestmeanstd{6.43}{1.08} & \sbestmeanstd{7.75}{1.02} & \sbestmeanstd{5.65}{1.19} \\
 & No predictive latent & \smeanstd{12.00}{3.30} & \smeanstd{13.31}{3.48} & \smeanstd{15.86}{3.84} & \smeanstd{19.09}{3.94} & \smeanstd{23.00}{4.26} & \smeanstd{16.65}{3.74} \\
 & FNO & \smeanstd{21.99}{0.65} & \smeanstd{21.49}{0.88} & \smeanstd{21.79}{1.10} & \smeanstd{22.64}{1.34} & \smeanstd{24.08}{1.60} & \smeanstd{22.40}{1.11} \\
\midrule
\multirow{3}{*}{Brusselator} & \modelname{} & \sbestmeanstd{3.79}{0.43} & \sbestmeanstd{4.85}{0.83} & \sbestmeanstd{6.18}{1.23} & \sbestmeanstd{7.61}{1.72} & \sbestmeanstd{9.90}{2.58} & \sbestmeanstd{6.47}{1.31} \\
 & No predictive latent & \smeanstd{9.52}{1.11} & \smeanstd{15.75}{1.65} & \smeanstd{21.94}{2.48} & \smeanstd{28.38}{3.66} & \smeanstd{34.40}{5.55} & \smeanstd{22.00}{2.85} \\
 & FNO & \smeanstd{9.08}{1.16} & \smeanstd{12.65}{1.11} & \smeanstd{17.39}{1.02} & \smeanstd{21.21}{1.27} & \smeanstd{23.87}{1.87} & \smeanstd{16.84}{0.86} \\
\midrule
\multirow{3}{*}{Ginzburg--Landau} & \modelname{} & \sbestmeanstd{9.38}{4.58} & \sbestmeanstd{10.94}{4.03} & \sbestmeanstd{12.90}{3.82} & \sbestmeanstd{14.83}{3.82} & \sbestmeanstd{16.88}{3.80} & \sbestmeanstd{12.99}{3.99} \\
 & No predictive latent & \smeanstd{21.73}{16.14} & \smeanstd{36.63}{25.34} & \smeanstd{49.09}{30.31} & \smeanstd{58.99}{31.61} & \smeanstd{67.52}{30.60} & \smeanstd{46.79}{26.74} \\
 & FNO & \smeanstd{25.62}{3.53} & \smeanstd{29.67}{4.96} & \smeanstd{38.39}{8.44} & \smeanstd{48.50}{12.08} & \smeanstd{58.40}{15.03} & \smeanstd{40.12}{8.69} \\
\midrule
\multirow{3}{*}{Schnakenberg} & \modelname{} & \sbestmeanstd{2.03}{0.42} & \sbestmeanstd{2.52}{0.47} & \sbestmeanstd{3.08}{0.64} & \sbestmeanstd{3.55}{0.81} & \sbestmeanstd{4.35}{1.08} & \sbestmeanstd{3.11}{0.68} \\
 & No predictive latent & \smeanstd{4.33}{0.77} & \smeanstd{7.08}{0.62} & \smeanstd{10.12}{0.67} & \smeanstd{13.26}{0.62} & \smeanstd{16.65}{0.73} & \smeanstd{10.29}{0.67} \\
 & FNO & \smeanstd{7.10}{0.20} & \smeanstd{8.62}{0.45} & \smeanstd{10.73}{0.56} & \smeanstd{12.72}{0.58} & \smeanstd{14.52}{0.53} & \smeanstd{10.74}{0.45} \\
\bottomrule
\end{tabular}%
}
\end{table}

\begin{table}[!htbp]
\centering
\scriptsize
\caption{{\textbf{Complete relative $L^2$ results on the five source systems at $K=10$.} Entries are $100\times$ the error and report the mean $\pm$ sample s.d. across three paired support-set selections. Within each run, errors are first averaged over all valid windows within each trajectory and then over the same 300 test trajectories (100 from each test regime). The Avg. column averages $h=1,\ldots,5$ within each run before across-run aggregation. Lower values are better. {Boldface marks the lowest numerical mean within each equation and horizon; no formal hypothesis tests were performed.}}}
\label{tab:supp_source_rel_k10}
\resizebox{\textwidth}{!}{%
\begin{tabular}{llcccccc}
\toprule
\textbf{PDE} & \textbf{Model} & \textbf{$h=1$} & \textbf{$h=2$} & \textbf{$h=3$} & \textbf{$h=4$} & \textbf{$h=5$} & \textbf{Avg.} \\
\midrule
\multirow{3}{*}{Gray--Scott} & \modelname{} & \sbestmeanstd{2.08}{0.88} & \sbestmeanstd{2.96}{1.07} & \sbestmeanstd{3.91}{1.25} & \sbestmeanstd{4.85}{1.38} & \sbestmeanstd{5.84}{1.39} & \sbestmeanstd{3.93}{1.19} \\
 & No predictive latent & \smeanstd{4.89}{2.13} & \smeanstd{7.95}{3.76} & \smeanstd{10.96}{4.89} & \smeanstd{13.84}{5.66} & \smeanstd{16.24}{5.94} & \smeanstd{10.78}{4.47} \\
 & FNO & \smeanstd{4.74}{0.86} & \smeanstd{5.77}{1.04} & \smeanstd{7.40}{1.17} & \smeanstd{9.13}{1.21} & \smeanstd{10.90}{1.20} & \smeanstd{7.59}{1.09} \\
\midrule
\multirow{3}{*}{FitzHugh--Nagumo} & \modelname{} & \sbestmeanstd{2.64}{0.56} & \sbestmeanstd{3.37}{0.59} & \sbestmeanstd{4.23}{0.80} & \sbestmeanstd{5.27}{1.04} & \sbestmeanstd{6.49}{1.17} & \sbestmeanstd{4.40}{0.83} \\
 & No predictive latent & \smeanstd{7.39}{1.78} & \smeanstd{9.19}{1.81} & \smeanstd{11.85}{2.27} & \smeanstd{14.93}{3.06} & \smeanstd{18.40}{3.70} & \smeanstd{12.35}{2.50} \\
 & FNO & \smeanstd{17.07}{2.42} & \smeanstd{16.65}{2.21} & \smeanstd{16.78}{2.00} & \smeanstd{17.33}{1.78} & \smeanstd{18.46}{1.58} & \smeanstd{17.26}{1.99} \\
\midrule
\multirow{3}{*}{Brusselator} & \modelname{} & \sbestmeanstd{2.33}{0.15} & \sbestmeanstd{3.14}{0.52} & \sbestmeanstd{3.98}{0.75} & \sbestmeanstd{4.77}{0.93} & \sbestmeanstd{6.19}{1.00} & \sbestmeanstd{4.08}{0.67} \\
 & No predictive latent & \smeanstd{3.19}{0.14} & \smeanstd{5.84}{0.06} & \smeanstd{8.41}{0.24} & \smeanstd{10.84}{0.41} & \smeanstd{13.20}{0.43} & \smeanstd{8.29}{0.20} \\
 & FNO & \smeanstd{6.46}{0.31} & \smeanstd{9.06}{0.16} & \smeanstd{12.15}{0.18} & \smeanstd{14.79}{0.31} & \smeanstd{17.05}{0.22} & \smeanstd{11.90}{0.15} \\
\midrule
\multirow{3}{*}{Ginzburg--Landau} & \modelname{} & \sbestmeanstd{4.87}{0.29} & \sbestmeanstd{6.42}{0.29} & \sbestmeanstd{8.22}{0.44} & \sbestmeanstd{10.02}{0.58} & \sbestmeanstd{11.99}{0.82} & \sbestmeanstd{8.30}{0.48} \\
 & No predictive latent & \smeanstd{11.10}{4.93} & \smeanstd{20.54}{10.28} & \smeanstd{28.69}{13.73} & \smeanstd{36.39}{16.22} & \smeanstd{44.66}{19.10} & \smeanstd{28.28}{12.85} \\
 & FNO & \smeanstd{18.52}{2.29} & \smeanstd{20.43}{1.61} & \smeanstd{26.50}{3.26} & \smeanstd{34.12}{4.84} & \smeanstd{41.99}{6.18} & \smeanstd{28.31}{3.25} \\
\midrule
\multirow{3}{*}{Schnakenberg} & \modelname{} & \sbestmeanstd{1.65}{0.03} & \sbestmeanstd{2.02}{0.01} & \sbestmeanstd{2.36}{0.05} & \sbestmeanstd{2.60}{0.10} & \sbestmeanstd{2.97}{0.32} & \sbestmeanstd{2.32}{0.09} \\
 & No predictive latent & \smeanstd{2.92}{0.50} & \smeanstd{4.76}{0.29} & \smeanstd{6.90}{0.41} & \smeanstd{9.15}{0.45} & \smeanstd{11.51}{0.50} & \smeanstd{7.05}{0.35} \\
 & FNO & \smeanstd{4.61}{0.50} & \smeanstd{6.26}{0.59} & \smeanstd{8.34}{0.64} & \smeanstd{10.36}{0.60} & \smeanstd{12.15}{0.56} & \smeanstd{8.34}{0.58} \\
\bottomrule
\end{tabular}%
}
\end{table}

\begin{table}[!htbp]
\centering
\scriptsize
\caption{{\textbf{Complete relative $L^2$ results on the five source systems at $K=20$.} Entries are $100\times$ the error and report the mean $\pm$ sample s.d. across three paired support-set selections. Within each run, errors are first averaged over all valid windows within each trajectory and then over the same 300 test trajectories (100 from each test regime). The Avg. column averages $h=1,\ldots,5$ within each run before across-run aggregation. Lower values are better. {Boldface marks the lowest numerical mean within each equation and horizon; no formal hypothesis tests were performed.}}}
\label{tab:supp_source_rel_k20}
\resizebox{\textwidth}{!}{%
\begin{tabular}{llcccccc}
\toprule
\textbf{PDE} & \textbf{Model} & \textbf{$h=1$} & \textbf{$h=2$} & \textbf{$h=3$} & \textbf{$h=4$} & \textbf{$h=5$} & \textbf{Avg.} \\
\midrule
\multirow{3}{*}{Gray--Scott} & \modelname{} & \sbestmeanstd{1.42}{0.04} & \sbestmeanstd{1.87}{0.09} & \sbestmeanstd{2.48}{0.18} & \sbestmeanstd{3.19}{0.26} & \sbestmeanstd{4.00}{0.36} & \sbestmeanstd{2.59}{0.17} \\
 & No predictive latent & \smeanstd{2.43}{0.24} & \smeanstd{3.97}{0.49} & \smeanstd{5.53}{0.75} & \smeanstd{7.19}{1.00} & \smeanstd{8.98}{1.15} & \smeanstd{5.62}{0.71} \\
 & FNO & \smeanstd{3.36}{0.53} & \smeanstd{4.24}{0.64} & \smeanstd{5.65}{0.72} & \smeanstd{7.25}{0.81} & \smeanstd{8.91}{0.87} & \smeanstd{5.88}{0.71} \\
\midrule
\multirow{3}{*}{FitzHugh--Nagumo} & \modelname{} & \sbestmeanstd{2.24}{0.32} & \sbestmeanstd{2.55}{0.08} & \sbestmeanstd{3.12}{0.11} & \sbestmeanstd{3.90}{0.13} & \sbestmeanstd{4.84}{0.15} & \sbestmeanstd{3.33}{0.11} \\
 & No predictive latent & \smeanstd{7.64}{0.40} & \smeanstd{8.74}{0.34} & \smeanstd{11.16}{0.61} & \smeanstd{13.96}{0.72} & \smeanstd{17.20}{0.58} & \smeanstd{11.74}{0.50} \\
 & FNO & \smeanstd{13.95}{1.43} & \smeanstd{13.61}{1.23} & \smeanstd{13.78}{0.99} & \smeanstd{14.52}{0.91} & \smeanstd{15.61}{0.79} & \smeanstd{14.29}{1.07} \\
\midrule
\multirow{3}{*}{Brusselator} & \modelname{} & \sbestmeanstd{2.01}{0.34} & \sbestmeanstd{2.35}{0.38} & \sbestmeanstd{2.67}{0.27} & \sbestmeanstd{3.07}{0.13} & \sbestmeanstd{4.21}{0.24} & \sbestmeanstd{2.86}{0.26} \\
 & No predictive latent & \smeanstd{3.00}{0.12} & \smeanstd{4.93}{0.15} & \smeanstd{7.09}{0.23} & \smeanstd{9.28}{0.30} & \smeanstd{11.56}{0.40} & \smeanstd{7.17}{0.23} \\
 & FNO & \smeanstd{4.65}{0.11} & \smeanstd{6.55}{0.09} & \smeanstd{8.78}{0.19} & \smeanstd{10.86}{0.16} & \smeanstd{12.70}{0.13} & \smeanstd{8.71}{0.12} \\
\midrule
\multirow{3}{*}{Ginzburg--Landau} & \modelname{} & \sbestmeanstd{4.08}{0.47} & \sbestmeanstd{5.32}{0.43} & \sbestmeanstd{6.80}{0.45} & \sbestmeanstd{8.39}{0.51} & \sbestmeanstd{10.24}{0.58} & \sbestmeanstd{6.97}{0.48} \\
 & No predictive latent & \smeanstd{10.81}{1.98} & \smeanstd{16.17}{2.55} & \smeanstd{20.34}{2.83} & \smeanstd{24.95}{3.66} & \smeanstd{30.28}{4.44} & \smeanstd{20.51}{2.93} \\
 & FNO & \smeanstd{12.26}{1.43} & \smeanstd{13.28}{1.20} & \smeanstd{16.61}{1.31} & \smeanstd{20.99}{1.56} & \smeanstd{25.87}{1.86} & \smeanstd{17.80}{1.46} \\
\midrule
\multirow{3}{*}{Schnakenberg} & \modelname{} & \sbestmeanstd{1.37}{0.27} & \sbestmeanstd{1.60}{0.26} & \sbestmeanstd{1.84}{0.31} & \sbestmeanstd{2.13}{0.38} & \sbestmeanstd{2.59}{0.25} & \sbestmeanstd{1.91}{0.29} \\
 & No predictive latent & \smeanstd{2.21}{0.20} & \smeanstd{3.67}{0.33} & \smeanstd{5.27}{0.34} & \smeanstd{6.94}{0.35} & \smeanstd{8.72}{0.42} & \smeanstd{5.36}{0.32} \\
 & FNO & \smeanstd{3.13}{0.30} & \smeanstd{4.33}{0.41} & \smeanstd{6.00}{0.50} & \smeanstd{7.74}{0.53} & \smeanstd{9.43}{0.49} & \smeanstd{6.12}{0.44} \\
\bottomrule
\end{tabular}%
}
\end{table}

\suppsubsection{S4.3 Errors in spatial first differences}
\label{supp:source_grad_results}

The mean absolute spatial first-difference errors defined in \cref{eq:supp_metric_gradient} are reported for $K=5$, $10$ and $20$ in \cref{tab:supp_source_grad_k5,tab:supp_source_grad_k10,tab:supp_source_grad_k20}, respectively. Each table gives the error at each forecast lead time and its average over $h=1,\ldots,5$.

\begin{table}[!htbp]
\centering
\scriptsize
\caption{{\textbf{Complete gradient $L^1$ results on the five source systems at $K=5$.} Entries are $100\times$ the error and report the mean $\pm$ sample s.d. across three paired support-set selections. Within each run, errors are first averaged over all valid windows within each trajectory and then over the same 300 test trajectories (100 from each test regime). The Avg. column averages $h=1,\ldots,5$ within each run before across-run aggregation. Lower values are better. {Boldface marks the lowest numerical mean within each equation and horizon; no formal hypothesis tests were performed.}}}
\label{tab:supp_source_grad_k5}
\resizebox{\textwidth}{!}{%
\begin{tabular}{llcccccc}
\toprule
\textbf{PDE} & \textbf{Model} & \textbf{$h=1$} & \textbf{$h=2$} & \textbf{$h=3$} & \textbf{$h=4$} & \textbf{$h=5$} & \textbf{Avg.} \\
\midrule
\multirow{3}{*}{Gray--Scott} & \modelname{} & \sbestmeanstd{0.35}{0.10} & \sbestmeanstd{0.53}{0.12} & \sbestmeanstd{0.70}{0.12} & \sbestmeanstd{0.86}{0.11} & \sbestmeanstd{1.02}{0.09} & \sbestmeanstd{0.69}{0.11} \\
 & No predictive latent & \smeanstd{0.72}{0.20} & \smeanstd{1.21}{0.32} & \smeanstd{1.63}{0.39} & \smeanstd{2.00}{0.41} & \smeanstd{2.32}{0.36} & \smeanstd{1.57}{0.33} \\
 & FNO & \smeanstd{0.69}{0.02} & \smeanstd{0.88}{0.03} & \smeanstd{1.16}{0.05} & \smeanstd{1.44}{0.06} & \smeanstd{1.72}{0.06} & \smeanstd{1.18}{0.04} \\
\midrule
\multirow{3}{*}{FitzHugh--Nagumo} & \modelname{} & \sbestmeanstd{0.16}{0.04} & \sbestmeanstd{0.19}{0.04} & \sbestmeanstd{0.23}{0.04} & \sbestmeanstd{0.28}{0.03} & \sbestmeanstd{0.34}{0.03} & \sbestmeanstd{0.24}{0.04} \\
 & No predictive latent & \smeanstd{0.44}{0.11} & \smeanstd{0.53}{0.12} & \smeanstd{0.65}{0.13} & \smeanstd{0.80}{0.14} & \smeanstd{0.96}{0.16} & \smeanstd{0.68}{0.13} \\
 & FNO & \smeanstd{1.16}{0.05} & \smeanstd{1.18}{0.05} & \smeanstd{1.23}{0.06} & \smeanstd{1.29}{0.06} & \smeanstd{1.38}{0.07} & \smeanstd{1.25}{0.06} \\
\midrule
\multirow{3}{*}{Brusselator} & \modelname{} & \sbestmeanstd{1.31}{0.16} & \sbestmeanstd{1.73}{0.13} & \sbestmeanstd{2.27}{0.26} & \sbestmeanstd{2.82}{0.39} & \sbestmeanstd{3.58}{0.55} & \sbestmeanstd{2.34}{0.25} \\
 & No predictive latent & \smeanstd{2.98}{0.26} & \smeanstd{5.02}{0.16} & \smeanstd{7.14}{0.22} & \smeanstd{9.15}{0.33} & \smeanstd{10.91}{0.54} & \smeanstd{7.04}{0.20} \\
 & FNO & \smeanstd{3.52}{0.36} & \smeanstd{4.51}{0.40} & \smeanstd{5.77}{0.32} & \smeanstd{6.74}{0.17} & \smeanstd{7.54}{0.15} & \smeanstd{5.62}{0.27} \\
\midrule
\multirow{3}{*}{Ginzburg--Landau} & \modelname{} & \sbestmeanstd{0.32}{0.11} & \sbestmeanstd{0.37}{0.08} & \sbestmeanstd{0.44}{0.07} & \sbestmeanstd{0.53}{0.07} & \sbestmeanstd{0.62}{0.06} & \sbestmeanstd{0.45}{0.08} \\
 & No predictive latent & \smeanstd{0.62}{0.42} & \smeanstd{1.05}{0.67} & \smeanstd{1.45}{0.85} & \smeanstd{1.81}{0.93} & \smeanstd{2.15}{0.95} & \smeanstd{1.42}{0.76} \\
 & FNO & \smeanstd{1.01}{0.07} & \smeanstd{1.10}{0.10} & \smeanstd{1.35}{0.19} & \smeanstd{1.66}{0.30} & \smeanstd{1.99}{0.39} & \smeanstd{1.42}{0.20} \\
\midrule
\multirow{3}{*}{Schnakenberg} & \modelname{} & \sbestmeanstd{0.31}{0.07} & \sbestmeanstd{0.36}{0.06} & \sbestmeanstd{0.44}{0.09} & \sbestmeanstd{0.51}{0.11} & \sbestmeanstd{0.64}{0.16} & \sbestmeanstd{0.45}{0.10} \\
 & No predictive latent & \smeanstd{0.60}{0.10} & \smeanstd{0.92}{0.08} & \smeanstd{1.29}{0.08} & \smeanstd{1.70}{0.07} & \smeanstd{2.14}{0.08} & \smeanstd{1.33}{0.08} \\
 & FNO & \smeanstd{1.24}{0.02} & \smeanstd{1.40}{0.03} & \smeanstd{1.63}{0.04} & \smeanstd{1.85}{0.03} & \smeanstd{2.07}{0.02} & \smeanstd{1.64}{0.03} \\
\bottomrule
\end{tabular}%
}
\end{table}

\begin{table}[!htbp]
\centering
\scriptsize
\caption{{\textbf{Complete gradient $L^1$ results on the five source systems at $K=10$.} Entries are $100\times$ the error and report the mean $\pm$ sample s.d. across three paired support-set selections. Within each run, errors are first averaged over all valid windows within each trajectory and then over the same 300 test trajectories (100 from each test regime). The Avg. column averages $h=1,\ldots,5$ within each run before across-run aggregation. Lower values are better. {Boldface marks the lowest numerical mean within each equation and horizon; no formal hypothesis tests were performed.}}}
\label{tab:supp_source_grad_k10}
\resizebox{\textwidth}{!}{%
\begin{tabular}{llcccccc}
\toprule
\textbf{PDE} & \textbf{Model} & \textbf{$h=1$} & \textbf{$h=2$} & \textbf{$h=3$} & \textbf{$h=4$} & \textbf{$h=5$} & \textbf{Avg.} \\
\midrule
\multirow{3}{*}{Gray--Scott} & \modelname{} & \sbestmeanstd{0.21}{0.07} & \sbestmeanstd{0.35}{0.11} & \sbestmeanstd{0.49}{0.14} & \sbestmeanstd{0.63}{0.17} & \sbestmeanstd{0.77}{0.17} & \sbestmeanstd{0.49}{0.13} \\
 & No predictive latent & \smeanstd{0.52}{0.25} & \smeanstd{0.93}{0.47} & \smeanstd{1.33}{0.62} & \smeanstd{1.68}{0.72} & \smeanstd{1.99}{0.74} & \smeanstd{1.29}{0.56} \\
 & FNO & \smeanstd{0.53}{0.03} & \smeanstd{0.70}{0.08} & \smeanstd{0.96}{0.10} & \smeanstd{1.23}{0.12} & \smeanstd{1.50}{0.12} & \smeanstd{0.98}{0.09} \\
\midrule
\multirow{3}{*}{FitzHugh--Nagumo} & \modelname{} & \sbestmeanstd{0.12}{0.02} & \sbestmeanstd{0.15}{0.02} & \sbestmeanstd{0.19}{0.02} & \sbestmeanstd{0.24}{0.03} & \sbestmeanstd{0.29}{0.03} & \sbestmeanstd{0.20}{0.03} \\
 & No predictive latent & \smeanstd{0.33}{0.03} & \smeanstd{0.41}{0.04} & \smeanstd{0.53}{0.06} & \smeanstd{0.66}{0.08} & \smeanstd{0.80}{0.10} & \smeanstd{0.55}{0.06} \\
 & FNO & \smeanstd{0.86}{0.04} & \smeanstd{0.87}{0.03} & \smeanstd{0.90}{0.02} & \smeanstd{0.95}{0.02} & \smeanstd{1.02}{0.02} & \smeanstd{0.92}{0.02} \\
\midrule
\multirow{3}{*}{Brusselator} & \modelname{} & \sbestmeanstd{0.96}{0.06} & \sbestmeanstd{1.29}{0.07} & \sbestmeanstd{1.64}{0.13} & \sbestmeanstd{1.98}{0.14} & \sbestmeanstd{2.51}{0.13} & \sbestmeanstd{1.67}{0.09} \\
 & No predictive latent & \smeanstd{1.22}{0.14} & \smeanstd{2.11}{0.15} & \smeanstd{2.98}{0.16} & \smeanstd{3.80}{0.18} & \smeanstd{4.57}{0.13} & \smeanstd{2.93}{0.14} \\
 & FNO & \smeanstd{2.67}{0.23} & \smeanstd{3.52}{0.21} & \smeanstd{4.48}{0.21} & \smeanstd{5.32}{0.22} & \smeanstd{6.08}{0.23} & \smeanstd{4.41}{0.22} \\
\midrule
\multirow{3}{*}{Ginzburg--Landau} & \modelname{} & \sbestmeanstd{0.20}{0.01} & \sbestmeanstd{0.26}{0.01} & \sbestmeanstd{0.34}{0.01} & \sbestmeanstd{0.42}{0.02} & \sbestmeanstd{0.51}{0.03} & \sbestmeanstd{0.35}{0.02} \\
 & No predictive latent & \smeanstd{0.38}{0.14} & \smeanstd{0.68}{0.30} & \smeanstd{0.95}{0.41} & \smeanstd{1.23}{0.50} & \smeanstd{1.56}{0.62} & \smeanstd{0.96}{0.39} \\
 & FNO & \smeanstd{0.82}{0.07} & \smeanstd{0.85}{0.05} & \smeanstd{1.03}{0.09} & \smeanstd{1.29}{0.15} & \smeanstd{1.57}{0.21} & \smeanstd{1.11}{0.10} \\
\midrule
\multirow{3}{*}{Schnakenberg} & \modelname{} & \sbestmeanstd{0.25}{0.00} & \sbestmeanstd{0.28}{0.00} & \sbestmeanstd{0.32}{0.00} & \sbestmeanstd{0.36}{0.00} & \sbestmeanstd{0.42}{0.03} & \sbestmeanstd{0.32}{0.00} \\
 & No predictive latent & \smeanstd{0.40}{0.08} & \smeanstd{0.63}{0.05} & \smeanstd{0.90}{0.06} & \smeanstd{1.18}{0.06} & \smeanstd{1.49}{0.06} & \smeanstd{0.92}{0.05} \\
 & FNO & \smeanstd{0.83}{0.09} & \smeanstd{1.00}{0.09} & \smeanstd{1.23}{0.08} & \smeanstd{1.46}{0.08} & \smeanstd{1.67}{0.07} & \smeanstd{1.24}{0.08} \\
\bottomrule
\end{tabular}%
}
\end{table}

\begin{table}[!htbp]
\centering
\scriptsize
\caption{{\textbf{Complete gradient $L^1$ results on the five source systems at $K=20$.} Entries are $100\times$ the error and report the mean $\pm$ sample s.d. across three paired support-set selections. Within each run, errors are first averaged over all valid windows within each trajectory and then over the same 300 test trajectories (100 from each test regime). The Avg. column averages $h=1,\ldots,5$ within each run before across-run aggregation. Lower values are better. {Boldface marks the lowest numerical mean within each equation and horizon; no formal hypothesis tests were performed.}}}
\label{tab:supp_source_grad_k20}
\resizebox{\textwidth}{!}{%
\begin{tabular}{llcccccc}
\toprule
\textbf{PDE} & \textbf{Model} & \textbf{$h=1$} & \textbf{$h=2$} & \textbf{$h=3$} & \textbf{$h=4$} & \textbf{$h=5$} & \textbf{Avg.} \\
\midrule
\multirow{3}{*}{Gray--Scott} & \modelname{} & \sbestmeanstd{0.16}{0.03} & \sbestmeanstd{0.23}{0.03} & \sbestmeanstd{0.32}{0.04} & \sbestmeanstd{0.41}{0.05} & \sbestmeanstd{0.52}{0.06} & \sbestmeanstd{0.33}{0.04} \\
 & No predictive latent & \smeanstd{0.29}{0.02} & \smeanstd{0.48}{0.07} & \smeanstd{0.69}{0.10} & \smeanstd{0.92}{0.13} & \smeanstd{1.18}{0.16} & \smeanstd{0.71}{0.10} \\
 & FNO & \smeanstd{0.40}{0.02} & \smeanstd{0.53}{0.04} & \smeanstd{0.76}{0.07} & \smeanstd{1.01}{0.10} & \smeanstd{1.27}{0.11} & \smeanstd{0.80}{0.07} \\
\midrule
\multirow{3}{*}{FitzHugh--Nagumo} & \modelname{} & \sbestmeanstd{0.10}{0.01} & \sbestmeanstd{0.12}{0.00} & \sbestmeanstd{0.15}{0.01} & \sbestmeanstd{0.19}{0.01} & \sbestmeanstd{0.24}{0.01} & \sbestmeanstd{0.16}{0.01} \\
 & No predictive latent & \smeanstd{0.35}{0.02} & \smeanstd{0.41}{0.01} & \smeanstd{0.51}{0.02} & \smeanstd{0.63}{0.03} & \smeanstd{0.76}{0.03} & \smeanstd{0.53}{0.02} \\
 & FNO & \smeanstd{0.67}{0.03} & \smeanstd{0.67}{0.03} & \smeanstd{0.70}{0.03} & \smeanstd{0.74}{0.03} & \smeanstd{0.80}{0.03} & \smeanstd{0.72}{0.03} \\
\midrule
\multirow{3}{*}{Brusselator} & \modelname{} & \sbestmeanstd{0.80}{0.04} & \sbestmeanstd{1.01}{0.11} & \sbestmeanstd{1.20}{0.07} & \sbestmeanstd{1.40}{0.02} & \sbestmeanstd{1.85}{0.05} & \sbestmeanstd{1.25}{0.05} \\
 & No predictive latent & \smeanstd{1.10}{0.04} & \smeanstd{1.84}{0.03} & \smeanstd{2.58}{0.06} & \smeanstd{3.31}{0.07} & \smeanstd{4.07}{0.08} & \smeanstd{2.58}{0.04} \\
 & FNO & \smeanstd{2.09}{0.11} & \smeanstd{2.69}{0.15} & \smeanstd{3.42}{0.19} & \smeanstd{4.12}{0.18} & \smeanstd{4.76}{0.17} & \smeanstd{3.41}{0.16} \\
\midrule
\multirow{3}{*}{Ginzburg--Landau} & \modelname{} & \sbestmeanstd{0.16}{0.01} & \sbestmeanstd{0.22}{0.02} & \sbestmeanstd{0.28}{0.03} & \sbestmeanstd{0.35}{0.04} & \sbestmeanstd{0.44}{0.05} & \sbestmeanstd{0.29}{0.03} \\
 & No predictive latent & \smeanstd{0.36}{0.07} & \smeanstd{0.53}{0.08} & \smeanstd{0.69}{0.08} & \smeanstd{0.87}{0.10} & \smeanstd{1.09}{0.11} & \smeanstd{0.71}{0.08} \\
 & FNO & \smeanstd{0.58}{0.05} & \smeanstd{0.60}{0.04} & \smeanstd{0.71}{0.05} & \smeanstd{0.88}{0.06} & \smeanstd{1.07}{0.07} & \smeanstd{0.77}{0.06} \\
\midrule
\multirow{3}{*}{Schnakenberg} & \modelname{} & \sbestmeanstd{0.22}{0.04} & \sbestmeanstd{0.23}{0.04} & \sbestmeanstd{0.26}{0.04} & \sbestmeanstd{0.31}{0.05} & \sbestmeanstd{0.39}{0.04} & \sbestmeanstd{0.28}{0.04} \\
 & No predictive latent & \smeanstd{0.30}{0.02} & \smeanstd{0.48}{0.02} & \smeanstd{0.69}{0.02} & \smeanstd{0.91}{0.02} & \smeanstd{1.14}{0.03} & \smeanstd{0.70}{0.02} \\
 & FNO & \smeanstd{0.60}{0.04} & \smeanstd{0.72}{0.05} & \smeanstd{0.90}{0.06} & \smeanstd{1.11}{0.06} & \smeanstd{1.32}{0.06} & \smeanstd{0.93}{0.05} \\
\bottomrule
\end{tabular}%
}
\end{table}

\suppsubsection{S4.4 Split-resolved forecasting results}
\label{supp:source_split_results}

{The main source-system analysis pools the in-distribution, coefficient-OOD and initial-condition-OOD trajectories. \Cref{tab:supp_source_split_summary} reports each split separately after equal-weight averaging over the five equations and five horizons. The three run-level values underlying every split summary are available in Supplementary Data~1.}

\begin{table}[!htbp]
\centering
\scriptsize
\caption{{\textbf{Split-resolved source-system forecasting.} Each entry reports the mean $\pm$ sample s.d. across three paired support-set selections. Within each run, errors are averaged over the indicated 100-trajectory split and then given equal weight across the five source equations and five forecast horizons. Values are multiplied by 100; lower values are better. {Boldface marks the lowest numerical mean within each split and adaptation budget; no formal hypothesis tests were performed.}}}
\label{tab:supp_source_split_summary}
\resizebox{\textwidth}{!}{%
\begin{tabular}{llcccccc}
\toprule
& & \multicolumn{3}{c}{\textbf{Relative $L^2$}} & \multicolumn{3}{c}{\textbf{Gradient $L^1$}} \\
\cmidrule(lr){3-5}\cmidrule(lr){6-8}
\textbf{Model} & \textbf{$K$} & \textbf{ID} & \textbf{Coeff.-OOD} & \textbf{IC-OOD} & \textbf{ID} & \textbf{Coeff.-OOD} & \textbf{IC-OOD} \\
\midrule
\multirow{3}{*}{\modelname{}} & 5 & $\mathbf{6.61 \pm 1.06}$ & $\mathbf{6.89 \pm 1.14}$ & $\mathbf{6.88 \pm 1.23}$ & $\mathbf{0.79 \pm 0.10}$ & $\mathbf{0.85 \pm 0.10}$ & $\mathbf{0.86 \pm 0.09}$ \\
 & 10 & $\mathbf{4.69 \pm 0.36}$ & $\mathbf{4.87 \pm 0.45}$ & $\mathbf{4.84 \pm 0.37}$ & $\mathbf{0.60 \pm 0.07}$ & $\mathbf{0.64 \pm 0.09}$ & $\mathbf{0.65 \pm 0.07}$ \\
 & 20 & $\mathbf{3.44 \pm 0.10}$ & $\mathbf{3.52 \pm 0.11}$ & $\mathbf{3.64 \pm 0.11}$ & $\mathbf{0.45 \pm 0.02}$ & $\mathbf{0.46 \pm 0.01}$ & $\mathbf{0.48 \pm 0.01}$ \\
\midrule
\multirow{3}{*}{No predictive latent} & 5 & $20.68 \pm 4.55$ & $21.73 \pm 5.34$ & $22.88 \pm 6.04$ & $2.29 \pm 0.15$ & $2.41 \pm 0.14$ & $2.51 \pm 0.19$ \\
 & 10 & $12.48 \pm 1.96$ & $13.28 \pm 2.81$ & $14.28 \pm 4.72$ & $1.27 \pm 0.14$ & $1.34 \pm 0.21$ & $1.38 \pm 0.21$ \\
 & 20 & $9.11 \pm 0.07$ & $9.65 \pm 0.23$ & $11.48 \pm 1.63$ & $1.00 \pm 0.00$ & $1.02 \pm 0.05$ & $1.12 \pm 0.06$ \\
\midrule
\multirow{3}{*}{FNO} & 5 & $19.98 \pm 1.22$ & $19.82 \pm 1.54$ & $20.02 \pm 1.49$ & $2.18 \pm 0.06$ & $2.19 \pm 0.07$ & $2.29 \pm 0.06$ \\
 & 10 & $14.52 \pm 0.40$ & $14.62 \pm 0.47$ & $14.90 \pm 0.77$ & $1.69 \pm 0.06$ & $1.70 \pm 0.07$ & $1.81 \pm 0.07$ \\
 & 20 & $10.35 \pm 0.39$ & $10.51 \pm 0.23$ & $10.83 \pm 0.24$ & $1.29 \pm 0.02$ & $1.29 \pm 0.03$ & $1.39 \pm 0.02$ \\
\bottomrule
\end{tabular}%
}
\end{table}

\suppnote{Supplementary Note 5: {Paired predictive-latent control on held-out governing equations}}
\label{supp:note_heldout_results}

\suppsubsection{S5.1 {Evaluation coverage, data mapping and interpretation}}
\label{supp:heldout_result_aggregation}

{The equation-level transfer evaluation covers Lambda--Omega, Barkley and Oregonator, $K\in\{1,5,10\}$ and direct forecast horizons $h=1,\ldots,5$. Figure~\ref{fig:ood_quantitative} reports the horizon-resolved comparison among \modelname{}, the independently trained no-predictive-latent control and FNO; the three run-level values underlying those curves are supplied as Supplementary Data~2. The broader horizon-averaged comparison with FNO, LNO, RieszNO, ReViT and CNextU-Net is already reported in main-text Table~\ref{tab:operator_comparison}, with its run-level values in Supplementary Data~3. To avoid duplicating those main-text display items, this note reports only the additional paired horizon-averaged comparison between \modelname{} and the no-predictive-latent control.}

{Within each support-set selection, the two models receive the same adaptation trajectories and are evaluated on the same 300 test trajectories. The control is retrained independently: it removes the trajectory-specific JEPA encoder--predictor pathway, retains the dense decoder and physical-context pathway, and supplies only spatially shared lead-time conditioning at the latent interface. It is therefore not an inference-time zeroing intervention. The comparison evaluates the contribution of the complete predictive-latent representation pathway under matched support sets and test trajectories.}

\suppsubsection{S5.2 {Horizon-averaged paired comparison}}
\label{supp:heldout_predictive_latent_control}

{For each equation, model, adaptation budget and support-set selection, errors are first averaged over all valid windows within each trajectory, then over the same 300 test trajectories and finally with equal weight over $h=1,\ldots,5$. The displayed error entries are the mean and sample standard deviation across the three resulting run-level values. Percentage reductions are computed within each paired support-set selection using Eq.~\eqref{eq:supp_paired_reduction} and only then summarized across the three paired reductions. Positive reductions indicate lower error for \modelname{}. Both the error summaries and paired reductions can be recomputed from the matched horizon-resolved run-level errors supplied as Supplementary Data~2.}

\begin{table}[!htbp]
\centering
\scriptsize
\caption{{\textbf{Paired comparison with the no-predictive-latent control on governing equations excluded from pretraining.} Error entries are $100\times$ the error and report the mean $\pm$ sample s.d. across three paired support-set selections. Within each selection, errors are first averaged over the same 300 test trajectories and then with equal weight over the five direct forecast horizons. Reduction entries are percentages computed within each paired selection as $100(1-E_{\mathrm{RD\mbox{-}JEPA}}/E_{\mathrm{control}})$ and then summarized across the three paired values. Positive reductions favour RD-JEPA. Lower error values are better; boldface marks the lower numerical mean in each paired comparison; no formal hypothesis tests were performed.}}
\label{tab:supp_heldout_predictive_latent_control}
\resizebox{\textwidth}{!}{%
\begin{tabular}{llccc@{\hspace{8pt}}ccc}
\toprule
& &
\multicolumn{3}{c}{\textbf{Relative $L^2$}} &
\multicolumn{3}{c}{\textbf{Gradient $L^1$}} \\
\cmidrule(lr){3-5}\cmidrule(lr){6-8}
\textbf{PDE} & \textbf{$K$} &
\textbf{RD-JEPA} & \textbf{No predictive latent} & \textbf{Reduction (\%)} &
\textbf{RD-JEPA} & \textbf{No predictive latent} & \textbf{Reduction (\%)} \\
\midrule

\multirow{3}{*}{Lambda--Omega}
 & 1
 & $\mathbf{12.24 \pm 2.58}$
 & $23.06 \pm 5.28$
 & $46.78 \pm 1.22$
 & $\mathbf{0.68 \pm 0.12}$
 & $1.24 \pm 0.28$
 & $45.11 \pm 2.17$ \\

 & 5
 & $\mathbf{9.85 \pm 0.79}$
 & $18.73 \pm 1.32$
 & $47.38 \pm 3.42$
 & $\mathbf{0.52 \pm 0.05}$
 & $0.97 \pm 0.05$
 & $46.39 \pm 4.69$ \\

 & 10
 & $\mathbf{8.38 \pm 1.46}$
 & $17.84 \pm 1.68$
 & $53.16 \pm 5.46$
 & $\mathbf{0.45 \pm 0.08}$
 & $0.94 \pm 0.10$
 & $52.01 \pm 4.76$ \\

\midrule

\multirow{3}{*}{Barkley}
 & 1
 & $\mathbf{10.53 \pm 2.55}$
 & $18.97 \pm 4.04$
 & $44.72 \pm 3.04$
 & $\mathbf{1.09 \pm 0.27}$
 & $1.97 \pm 0.42$
 & $44.60 \pm 2.64$ \\

 & 5
 & $\mathbf{6.90 \pm 0.24}$
 & $12.97 \pm 0.85$
 & $46.76 \pm 1.76$
 & $\mathbf{0.74 \pm 0.04}$
 & $1.37 \pm 0.12$
 & $46.35 \pm 2.46$ \\

 & 10
 & $\mathbf{5.23 \pm 0.32}$
 & $10.06 \pm 0.67$
 & $47.95 \pm 1.01$
 & $\mathbf{0.58 \pm 0.04}$
 & $1.10 \pm 0.05$
 & $47.35 \pm 0.74$ \\

\midrule

\multirow{3}{*}{Oregonator}
 & 1
 & $\mathbf{11.36 \pm 1.34}$
 & $22.19 \pm 3.93$
 & $48.16 \pm 8.02$
 & $\mathbf{0.47 \pm 0.05}$
 & $1.01 \pm 0.12$
 & $53.46 \pm 3.77$ \\

 & 5
 & $\mathbf{4.83 \pm 0.79}$
 & $9.86 \pm 2.45$
 & $50.32 \pm 4.86$
 & $\mathbf{0.26 \pm 0.06}$
 & $0.53 \pm 0.16$
 & $50.50 \pm 2.94$ \\

 & 10
 & $\mathbf{4.05 \pm 0.47}$
 & $8.06 \pm 1.04$
 & $49.63 \pm 0.76$
 & $\mathbf{0.22 \pm 0.02}$
 & $0.45 \pm 0.03$
 & $52.11 \pm 0.58$ \\

\bottomrule
\end{tabular}%
}
\end{table}

\suppnote{Supplementary Note 6: {Matched control and boundary-condition stress test}}
\label{supp:note_robustness}

\suppsubsection{S6.1 {Architecture-matched training without predictive pretraining}}
\label{supp:scratch_control_results}

{The architecture-matched from-scratch control is evaluated only on the three governing equations excluded from pretraining. For every equation and value of $K$, RD-JEPA and the control receive the same support trajectories and are evaluated on the same 300 test trajectories. \Cref{tab:supp_scratch_control} reports horizon-averaged results.}

Across all three target equations and all three adaptation budgets, RD-JEPA has lower mean relative $L^2$ and gradient $L^1$ errors than the architecture-matched scratch model (Supplementary Table~\ref{tab:supp_scratch_control}). Under matched support sets and update budgets, this ordering shows that source pretraining supplies reusable information that the same architecture does not recover from the limited target trajectories alone. The three run-level values underlying each entry are provided in Supplementary Data~4.

\begin{table}[!htbp]
\centering
\scriptsize
\caption{{\textbf{Architecture-matched control trained without predictive pretraining on governing equations excluded from pretraining.} Each entry is $100\times$ the error and reports the mean $\pm$ sample s.d. across three paired support-set selections. For each selection, errors are first averaged over the same 300 test trajectories and then over the five direct forecast horizons. The model-training random-number-generator state is fixed, and both models use the same support trajectories within each selection. Lower values are better. Boldface marks the lower numerical mean in each paired comparison; no formal hypothesis tests were performed.}}
\label{tab:supp_scratch_control}
\resizebox{\textwidth}{!}{%
\begin{tabular}{llcccccc}
\toprule
& & \multicolumn{3}{c}{\textbf{Relative $L^2$ $\downarrow$ $(\times10^{-2})$}} &
\multicolumn{3}{c}{\textbf{Gradient $L^1$ $\downarrow$ $(\times10^{-2})$}} \\
\cmidrule(lr){3-5}\cmidrule(lr){6-8}
\textbf{PDE} & \textbf{Model} &
\textbf{$K=1$} & \textbf{$K=5$} & \textbf{$K=10$} &
\textbf{$K=1$} & \textbf{$K=5$} & \textbf{$K=10$} \\
\midrule
\multirow{2}{*}{Lambda--Omega}
 & \modelname{} & $\mathbf{12.24 \pm 2.58}$ & $\mathbf{9.85 \pm 0.79}$ & $\mathbf{8.38 \pm 1.46}$ & $\mathbf{0.68 \pm 0.12}$ & $\mathbf{0.52 \pm 0.05}$ & $\mathbf{0.45 \pm 0.08}$ \\
 & Architecture-matched scratch & $23.42 \pm 4.73$ & $20.14 \pm 0.91$ & $19.11 \pm 1.79$ & $1.27 \pm 0.26$ & $1.05 \pm 0.04$ & $1.00 \pm 0.12$ \\
\midrule
\multirow{2}{*}{Barkley}
 & \modelname{} & $\mathbf{10.53 \pm 2.55}$ & $\mathbf{6.90 \pm 0.24}$ & $\mathbf{5.23 \pm 0.32}$ & $\mathbf{1.09 \pm 0.27}$ & $\mathbf{0.74 \pm 0.04}$ & $\mathbf{0.58 \pm 0.04}$ \\
 & Architecture-matched scratch & $19.23 \pm 3.48$ & $13.03 \pm 0.48$ & $10.01 \pm 1.12$ & $2.01 \pm 0.38$ & $1.39 \pm 0.06$ & $1.10 \pm 0.12$ \\
\midrule
\multirow{2}{*}{Oregonator}
 & \modelname{} & $\mathbf{11.36 \pm 1.34}$ & $\mathbf{4.83 \pm 0.79}$ & $\mathbf{4.05 \pm 0.47}$ & $\mathbf{0.47 \pm 0.05}$ & $\mathbf{0.26 \pm 0.06}$ & $\mathbf{0.22 \pm 0.02}$ \\
 & Architecture-matched scratch & $22.12 \pm 2.82$ & $9.43 \pm 1.89$ & $7.37 \pm 1.00$ & $0.90 \pm 0.08$ & $0.51 \pm 0.12$ & $0.39 \pm 0.03$ \\
\bottomrule
\end{tabular}%
}
\end{table}

\suppsubsection{S6.2 {Few-trajectory adaptation under non-periodic boundary conditions}}
\label{supp:boundary_shift_results}

We next assess whether a representation pretrained exclusively on periodic systems remains useful after adaptation to Barkley dynamics with homogeneous Neumann, Robin or Dirichlet boundary conditions. RD-JEPA, the no-predictive-latent control and FNO are trained separately for each boundary condition and $K\in\{1,5,10\}$. All models use the same support trajectories within each of the three support-set selections and are evaluated on the same 300 boundary-specific test trajectories. The RD-JEPA predictor retains its pretrained periodic latent-neighbour map, and no model receives an explicit boundary-condition label.

RD-JEPA has the lowest mean relative $L^2$ and gradient $L^1$ error in all nine boundary-condition--budget settings (Supplementary Table~\ref{tab:supp_boundary_shift}). The ordering holds for homogeneous Neumann, Robin and Dirichlet conditions and for $K=1$, $5$ and $10$. These results extend the observed transfer to a combined boundary-condition and discretization shift within Barkley at the evaluated equation, domain, resolution and simulator. The three run-level values underlying each entry are provided in Supplementary Data~5.

\begin{table}[!htbp]
\centering
\scriptsize
\caption{{
\textbf{Few-trajectory Barkley forecasting under non-periodic boundary conditions.}
Each entry is $100\times$ the error and reports the mean $\pm$ sample s.d.
across three paired support-set selections.
For each selection, errors are first averaged over the same 300 test
trajectories and then over the five direct forecast horizons.
The model-training random-number-generator state is fixed, and all models
use the same support trajectories within each selection.
Lower values are better. Boldface marks the lowest mean for each boundary
condition, metric and adaptation budget.}}
\label{tab:supp_boundary_shift}
\resizebox{\textwidth}{!}{%
\begin{tabular}{llcccccc}
\toprule
& & \multicolumn{3}{c}{\textbf{Relative $L^2$ $\downarrow$ $(\times10^{-2})$}} &
\multicolumn{3}{c}{\textbf{Gradient $L^1$ $\downarrow$ $(\times10^{-2})$}} \\
\cmidrule(lr){3-5}\cmidrule(lr){6-8}
\textbf{Boundary condition} & \textbf{Model} &
\textbf{$K=1$} & \textbf{$K=5$} & \textbf{$K=10$} &
\textbf{$K=1$} & \textbf{$K=5$} & \textbf{$K=10$} \\
\midrule

\multirow{3}{*}{Neumann}
 & \modelname{}
 & $\mathbf{10.79 \pm 1.42}$
 & $\mathbf{7.33 \pm 1.30}$
 & $\mathbf{5.49 \pm 0.46}$
 & $\mathbf{0.68 \pm 0.10}$
 & $\mathbf{0.48 \pm 0.08}$
 & $\mathbf{0.38 \pm 0.03}$ \\
 & No predictive latent
 & $26.88 \pm 5.58$
 & $17.78 \pm 3.24$
 & $10.83 \pm 0.91$
 & $1.69 \pm 0.33$
 & $1.16 \pm 0.18$
 & $0.75 \pm 0.09$ \\
 & FNO
 & $30.89 \pm 1.50$
 & $19.15 \pm 2.32$
 & $15.02 \pm 0.76$
 & $2.02 \pm 0.09$
 & $1.21 \pm 0.13$
 & $0.97 \pm 0.04$ \\

\midrule
\multirow{3}{*}{Robin}
 & \modelname{}
 & $\mathbf{11.13 \pm 1.35}$
 & $\mathbf{7.21 \pm 1.26}$
 & $\mathbf{5.57 \pm 0.39}$
 & $\mathbf{0.69 \pm 0.10}$
 & $\mathbf{0.47 \pm 0.08}$
 & $\mathbf{0.39 \pm 0.02}$ \\
 & No predictive latent
 & $25.61 \pm 4.46$
 & $17.91 \pm 3.13$
 & $10.79 \pm 2.67$
 & $1.61 \pm 0.28$
 & $1.18 \pm 0.17$
 & $0.76 \pm 0.19$ \\
 & FNO
 & $30.61 \pm 1.31$
 & $18.88 \pm 2.24$
 & $14.95 \pm 0.66$
 & $2.01 \pm 0.06$
 & $1.20 \pm 0.13$
 & $0.96 \pm 0.03$ \\

\midrule
\multirow{3}{*}{Dirichlet}
 & \modelname{}
 & $\mathbf{11.17 \pm 1.76}$
 & $\mathbf{7.48 \pm 1.95}$
 & $\mathbf{5.26 \pm 0.21}$
 & $\mathbf{0.70 \pm 0.10}$
 & $\mathbf{0.49 \pm 0.12}$
 & $\mathbf{0.36 \pm 0.00}$ \\
 & No predictive latent
 & $28.04 \pm 2.91$
 & $18.34 \pm 3.55$
 & $11.14 \pm 1.03$
 & $1.73 \pm 0.19$
 & $1.21 \pm 0.22$
 & $0.78 \pm 0.08$ \\
 & FNO
 & $31.76 \pm 1.00$
 & $19.50 \pm 2.41$
 & $15.29 \pm 0.92$
 & $2.04 \pm 0.07$
 & $1.21 \pm 0.13$
 & $0.97 \pm 0.04$ \\

\bottomrule
\end{tabular}%
}
\end{table}

\end{document}